\documentclass{article}

\usepackage[preprint]{neurips_2026}
\setcitestyle{numbers,square}

\usepackage[utf8]{inputenc} 
\usepackage[T1]{fontenc}    
\usepackage{hyperref}       
\usepackage{url}            
\usepackage{booktabs}       
\usepackage{amsfonts}       
\usepackage{nicefrac}       
\usepackage{microtype}      
\usepackage{xcolor}         
\usepackage{amsmath}
\usepackage{comment}
\usepackage{graphicx}
\usepackage{subcaption}

\newcommand{\modelname}[1]{\textcolor{black}{{{LC-CBM}}}}

\makeatletter
\renewcommand{\@noticestring}{%
  Preprint. Work in progress; results are preliminary and may be revised
  in future versions.%
}
\makeatother

\usepackage{multirow}
\title{Auditing Where Concept Bottleneck Accuracy Comes From on CUB}

\author{%
  Sascha Saralajew\thanks{Equal contribution.}\\
  NEC Labs Europe\\
  Heidelberg, Germany\\
  \texttt{sascha.saralajew@neclab.eu} \\
  \And
  Eleonora Poeta$^{*}$\\
  Politecnico di Torino\\
  Turin, Italy\\
  \texttt{eleonora.poeta@polito.it} \\
  \And
  Eliana Pastor\\
  Politecnico di Torino\\
  Turin, Italy\\
  \texttt{eliana.pastor@polito.it} \\
  \And
  Ammar Shaker\\
  NEC Labs Europe\\
  Heidelberg, Germany\\
  \texttt{ammar.shaker@neclab.eu} \\
}

\begin{document}

\maketitle

\begin{abstract}
Concept Bottleneck Models (CBMs) promise predictions that humans can
inspect and correct through concepts such as ``curved beak''. On
CUB-200-2011, their standard visual benchmark, progress is nevertheless
measured like that of ordinary classifiers: by higher task and concept
accuracy. Neither number reveals whether the label is predicted from
the exposed concept scores at all, or whether a predicted concept is
supported by visual evidence a human would recognize as that concept,
rather than by other image features that co-occur with it in the
training data, such as cues to the species as a whole. We audit where
the accuracy of CBMs and Concept Embedding Models (CEMs) on CUB comes
from. We first show that the CUB concept annotations are too noisy to
explain this accuracy: the class-level labels used for training agree
with the per-image annotations on only $\approx$82\% of image--concept
pairs, and classifiers trained on the per-image annotations reach only
$\approx$37\% species accuracy. Leakage probes show that CEM predictions cannot be reconstructed
from their exposed concept scores, with the missing class information
stored in internal embeddings. We identify four design axes through
which class information can enter concept scores---receptive field,
shared versus concept-specific representations, norm-dependent
scoring, and task supervision---and a systematic ablation shows that
most of the class information in concept scores enters through
non-local features:
restricting a ResNet50 backbone to local BagNet patches lowers the
accuracy recoverable from concept scores from 72\% to 43\%, close to
what the ground-truth concept annotations themselves support, while
joint task training raises it again at the expense of concept
accuracy. Spatial localization and part occlusion show that locality
makes grounding measurable but does not guarantee it: under
class-level labels, local concept scores use the head as a proxy for
body, wing, and tail concepts, and CEM's concept scores barely respond
when their supporting part is occluded. Task and concept accuracy
alone are therefore poor proxies for whether a CBM delivers the
interpretability and intervenability it promises: they should support,
not drive, the evaluation of architectural changes, and given the
fuzziness of the CUB annotations, differences in CUB accuracy should
not be overinterpreted.
\end{abstract}

\section{Introduction}
\label{sec:intro}
Concept Bottleneck Models (CBMs)~\cite{koh2020concept,poeta2023concept}
are designed to make predictions through human-understandable concepts,
such as ``has red wing'' or ``long beak''. Their appeal rests on two
promises beyond predictive accuracy: \emph{interpretability} -- a
decision can be inspected through the concepts that produced it -- and
\emph{intervenability} -- a decision can be corrected by correcting
those concepts.

The CUB-200-2011 dataset~\cite{wah2011caltech} is the standard visual
benchmark for this paradigm. As a result, progress in CBMs is often
summarized by higher classification accuracy on CUB, whether reported
at the class (species) level or at the instance level -- the same
metric used to evaluate ordinary classifiers. This creates a tension:
neither of these accuracy numbers confirms the interpretability
promise above -- that predictions are actually mediated by concepts a
human can recognize in the image, e.g., actually seeing a curved
beak, not merely a pattern that correlates with the label.

The interpretability promise can fail in two distinct ways. First, the
label predictor could rely on information that the concepts do not
express: either by bypassing the exposed concept scores -- the model's
\emph{concept interface} -- altogether (e.g., through the internal
concept embedding in CEMs), or through structured variation encoded
within the concept scores themselves, as soft concept scores are known
to do~\cite{mahinpei2021promises,havasi2022addressing}. Second, even
when the label predictor uses only the concept scores, those scores
could be predicted from image features that co-occur with the concept
in the training data but that a human would not perceive as evidence
for it -- e.g., predicting ``curved beak'' from cues to the species as
a whole. Such a concept score is predictive, but not visually grounded
in the sense a human inspecting the image would expect.

This second failure mode also undermines intervenability. The label
predictor learns what a concept score means from the model's own
scores: if the score for ``curved beak'' in fact tracks species cues
elsewhere in the image, the label predictor learns to read it as
species evidence. A human who corrects this score after inspecting the
beak then feeds the label predictor a value whose meaning differs from
the one it was trained on, so the effect of the correction on the
prediction no longer reflects the human's judgment about the beak. We
do not evaluate interventions directly; instead, we test their
precondition, namely whether concept scores are driven by the evidence
a human would inspect.

Both failure modes allow a model to be more accurate than its concepts
can explain. As a first check, we therefore estimate how much task
information the CUB concept annotations themselves provide, by
training classifiers on ground-truth instance-level concept vectors.
The resulting concept-only reference accuracy is only $\approx$37\%.
Because the annotations are noisy -- a single Amazon Mechanical Turk
annotator per image -- this value marks a reference band rather than a
strict upper bound on faithful concept reasoning -- yet even allowing
for this noise, it is far below the 70\%+ accuracies reported by
widely used CBM and CEM models, which suggests that concept
annotations alone do not explain those numbers.

This motivates our central question: where does the extra accuracy
come from, and is it consistent with the interpretability and
intervenability promises introduced above? We address the two failure
modes in turn. For the first, leakage probes test whether the native
predictions of CBMs and CEMs can be reconstructed from their exposed
concept scores alone. For the second, we ask how class information
enters the concept scores themselves. We identify four design choices
through which this can happen -- the receptive field of the features,
shared versus concept-specific representations, norm-dependent
scoring, and task supervision of the concept scores -- and sweep
receptive field, head sharing, and task supervision systematically
within one model template, measuring how much class
information a probe recovers from the frozen concept scores of each
configuration. The receptive field turns out to be the main lever, and
the configuration in which all four choices are constrained carries
roughly as much class information as the ground-truth concept
annotations. Because this configuration computes each concept score
from a local image patch, it also makes grounding measurable: spatial
localization tests whether its concept scores peak on the image regions
a human would use to recognize the concept. Part occlusion, which
applies to any model, tests whether concept scores respond when that
evidence is removed; we apply it to the constrained configuration and
to CEM, the model with the largest bypass gap.

\textbf{Contributions.}
Our contributions are fourfold:
(i) We show that the CUB concept annotations alone cannot account for
the accuracies reported by CBM/CEM models: classifiers trained
directly on ground-truth concept vectors reach only $\approx$37\% task
accuracy -- far below the 70\%+ typically reported -- and this
gap partly reflects a structural mismatch between the class-level
concept labels used for training (identical for every image of a
species) and the instance-level concept evidence actually visible in
each image;
(ii) We identify four design axes -- receptive field, shared concept
representations, norm-dependent scoring, and task supervision --
through which CBMs and CEMs can place class information in their
concept scores that is not backed by visual evidence for the annotated
concepts;
(iii) We audit where CBM and CEM accuracy on CUB comes from: leakage
probes show that CEM predictions partly bypass the exposed concept
scores, and a systematic ablation over these axes shows that
non-local features and task supervision are the main sources of class
information in concept scores, while closing all four pathways brings
this information down to the level of the concept annotations;
(iv) We show that locality makes grounding measurable but does not
guarantee it: under class-level supervision, local concept scores use
the head as a proxy for body, wing, and tail concepts, and CEM concept
scores barely respond when the evidence that should support them is
occluded.

\section{Related Work}
\label{sec:related}

\textbf{Concept models and the CUB benchmark.}
CBMs~\cite{koh2020concept} decompose prediction into a concept
predictor followed by a downstream label predictor, making the
intermediate concept vector available for inspection and intervention.
Subsequent models increase flexibility through continuous concept
embeddings~\cite{espinosa2022concept}, probabilistic concept
uncertainty~\cite{kim2023probabilistic}, and broader concept-based
architectures~\cite{poeta2023concept}. CUB-200-2011~\cite{wah2011caltech}
has become the standard visual benchmark for this line of work because
it provides bird species labels, attribute annotations, and part
locations. This makes it a natural testbed for CBMs, but species
accuracy on CUB is a weak signal of concept use: images of one species
share many visual cues beyond the annotated attributes, so a model can
recognize the species from those cues even when its concept scores are
not grounded in the attributes they name.

\textbf{Information leakage in concept models.}
Mahinpei et al.~\cite{mahinpei2021promises} and Havasi et
al.~\cite{havasi2022addressing} show that soft concept scores can
encode task information unavailable to hard concept predictors;
discretizing the scores~\cite{havasi2022addressing} or deriving them
from discrete prototypes~\cite{colamonaco2026prototype} is therefore
often taken to prevent leakage. Under class-level labels, however, a
hard concept vector can still identify the species
(Section~\ref{sec:ceilings}), so discreteness alone does not remove
class information from the concept interface. Parisini et
al.~\cite{parisini2025leakage} develop an information-theoretic account
of concept-task and interconcept leakage and show that CEM-style
representations can carry task information beyond scalar concept
activations. Related work proposes direct leakage metrics and
mitigations, including irrelevant-concept contribution
scores~\cite{schoen2025measuring} and information-theoretic leakage
measures~\cite{makonnen2025measuring}. Espinosa
Zarlenga~\cite{espinosa2026defense} argues that leakage is not always
harmful: when the concept set is incomplete, some leakage can be
necessary for accurate and intervenable models, provided it is
properly localized. We share this view and do not treat every gain in
accuracy as a defect. CUB is such a concept-incomplete setting, and
deciding whether its leakage is benign requires knowing where the
additional information comes from and whether the concept scores stay
grounded. Our audit provides exactly this decomposition, across exposed
concept scores, internal representations, receptive-field choices, and
supervision regimes.

\textbf{Evaluating concept-interface faithfulness.}
Intervention metrics~\cite{koh2020concept,espinosa2022concept},
concept-representation metrics~\cite{zarlenga2023towards}, and
completeness scores~\cite{yeh2020completeness} are common ways to
evaluate whether concepts are useful. These metrics answer different
questions from the one studied here. Interventions test whether user
corrections improve predictions; completeness compares concepts to a
reference predictor; leakage metrics quantify specific forms of
unintended information. We instead start from the benchmark itself:
given the CUB annotations and the exposed concept interface, how much
of the reported task accuracy is visible at that interface, and which
architectural choices make the remainder available?

\textbf{Spatial grounding and local evidence.}
For visual concepts, faithfulness also has a spatial component: a model
may predict a concept correctly while using evidence from the wrong
image region, a form of shortcut learning~\cite{geirhos2020shortcut}.
On CUB, Margeloiu et al.~\cite{margeloiu2021concept}
give qualitative saliency-map examples in which concepts such as wing
pattern or leg color attend to the whole bird rather than to the
corresponding part. Huang et al.~\cite{huang2024concept} turn this
observation into a benchmark: using the CUB part keypoints, they test
whether the peak of a concept's localization map falls near the
annotated part, and find that most concepts of standard CBMs fail this
test across eight backbones. Raman et al.~\cite{raman2025locality}
formalize the property as \emph{locality} and measure it with
perturbation, intervention, and masking metrics; their CUB masking
results for CEM and ProbCBM change concept accuracy by at most 3.5
points, whether the relevant or an irrelevant region is masked, and
Jeon et al.~\cite{jeon2025locality} find that the concept scores of
label-free CBMs retain 87--90\% of their value when the concept's
region is deleted. We read these results as
insensitivity of the concept predictions to the concept region, in
line with our part-occlusion results for CEM
(Section~\ref{sec:occlusion_faithfulness}).

Several methods aim to restore locality: part-prototype CBMs in the
style of ProtoPNet~\cite{chen2019looks} with feature-alignment
losses~\cite{huang2024concept}, CLIP-guided
prototypes with a localization loss~\cite{jeon2025locality}, and
concept prototypes computed on segmented image
parts~\cite{colamonaco2026prototype}. These approaches encourage
locality through training objectives or external segmenters rather
than guaranteeing it: deep feature maps need not be spatially aligned
with the input~\cite{hoffmann2021looks,huang2024concept}, and on CUB,
where no segmenter is used, PGCM treats the whole image as a single
part~\cite{colamonaco2026prototype}. Following BagNet-style
models~\cite{brendel2019approximating,djoumessi2024actually}, we
instead restrict the receptive field \emph{by construction}: each
concept score is computed from a single image patch of known size, so
locality holds exactly and the patch behind every score can be traced
without attribution methods. We use this restriction as a controlled
ablation axis to ask how much of the \emph{task accuracy} reported on
CUB depends on non-local evidence. Our localization and occlusion tests
build on the keypoint- and removal-based protocols
above~\cite{huang2024concept,raman2025locality}, but break results down
by anatomical part and compare class- with instance-level supervision.
This comparison shows that locality alone does not ensure grounding:
under class-level labels, local concept scores use the head as a proxy
for other parts.

\begin{figure}
    \centering
    \includegraphics[width=\linewidth]{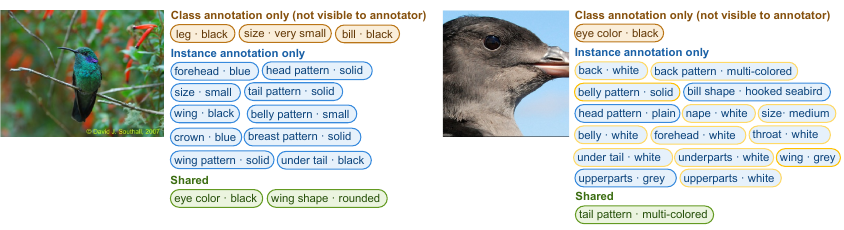}
    \caption{Comparison of class-level and instance-level annotations for two sample images. Concepts are color-coded by source: class-only (\textcolor[HTML]{B26B00}{orange}; not visually observable to the annotator), instance-only (\textcolor[HTML]{2F6FB0}{blue}), and shared (\textcolor[HTML]{4F7F2B}{green}).  In \textcolor[HTML]{F0E442}{yellow}, we highlight annotated concepts judged by the authors to be not visually observable. A subset of the concepts is shown for visualization purposes. }
    \label{fig:examples_two}
\end{figure}

\section{Preliminaries}
\label{sec:prelim}

\paragraph{Concept Bottleneck Models}
Concept-bottleneck models (CBMs)~\cite{espinosa2022concept,
koh2020concept} can be written as a composition
$f \circ s \circ g : \mathcal{X} \xrightarrow{g} \mathcal{Z}
\xrightarrow{s} \mathcal{C} \xrightarrow{f} \mathcal{Y}$,
where $\mathcal{X} \subseteq \mathbb{R}^D$ is the input space,
$\mathcal{Z}$ a latent space, $\mathcal{C} \subseteq [0,1]^k$ the
concept space, and $\mathcal{Y}$ the label space. Concept-based data
take the form $\{(x_i, c_i, y_i)\}_{i=1}^n$, where $c_i$ is the
concept annotation vector for $x_i$. The defining constraint---the
\emph{concept bottleneck}---is that the predicted class, $\hat{y}$, depends on $x$ only through the predicted concepts, $\hat{c}$, limiting the task-relevant information accessible to the classifier.

\paragraph{The CUB Benchmark and Its Annotations}
The Caltech-UCSD Birds (CUB) dataset~\cite{wah2011caltech} is \textit{the
principal benchmark} for CBMs. CUB contains 11{,}788 images across 200
bird species, each annotated with 312 visual attributes (concepts). In the
original dataset, annotations are at the \emph{instance level}: each
image is annotated by a single annotator who records per concept a binary label (present or absent), a
visibility indicator via a certainty score $\kappa \in \{1,2,3,4\}$
ranging from not observable ($1$) to definitely ($4$). CUB also provides
part keypoints for 15 anatomical regions with visibility flags, which
we use for spatial faithfulness evaluation.

In practice, CBM evaluations rely on a processed version introduced by
Koh et al.~\cite{koh2020concept}: 312 attributes are reduced to 112
binary concepts, and \emph{class-level} concept labels are obtained via
majority voting so that \textit{all} images of the same species share identical concept
labels. While intended to reduce noise, this introduces aggregation
artifacts. We refer to the two protocols as \emph{by-class}
(majority-voted) and \emph{by-instance} (original per-image
annotations).

\section{Annotation Limits and a Concept-Only Reference for CUB}
\label{sec:ceilings}
The premise of CBMs is that predictions are mediated by human-annotated concepts; a model that adheres strictly to this premise can only be as accurate as the concepts allow. We show that the CUB concept annotations~\cite{wah2011caltech}, both in their original form and as processed by prior work~\cite{koh2020concept}, limit how reliably concepts can be supervised and how much task information they can carry. 
The \textit{by-class} protocol~\cite{koh2020concept} assigns unique but identical concept vectors to all images of a species, meaning a model that identifies the correct species can achieve 100\% concept accuracy without inspecting any image-level evidence, simply by using the predicted class label as a key for per-class lookup table (see Appendix~\ref{app:concept_ceiling}). 
The \textit{by-instance} protocol~\cite{wah2011caltech} provides visually grounded supervision but introduces its own noise via \textit{single}-annotator labels and subjective certainty scores. Both regimes, therefore, impose a structural limit on performance.

\paragraph{By-Instance vs.\ By-Class Supervision}
Faithful concept-based reasoning requires attributes to be \emph{visible} in each image, making instance-level annotations the most appropriate supervision signal. To compare \textit{by-class} and \textit{by-instance} protocols, we measure their \textit{agreement} as the average concept-wise accuracy between instance-level annotations and class-level labels across all image--concept pairs, corresponding to an oracle that predicts true image-level concepts evaluated under class-level supervision.
Agreement is bounded at $\approx$82\% (at $\kappa \geq 1$) and increases only marginally to $\approx$84\% under high-certainty annotations $\kappa = 4$ (retaining $\approx$55\% of pairs; Appendix~\ref{app:concept_ann_cub_koh}, Fig.~\ref{fig:cub_annotation_mismatch}(a)), indicating a structural mismatch rather than one driven by low-confidence labels.
Concretely, for roughly one in five image--concept pairs, the class-level label and the annotator's judgment for that image disagree: either the class-level label marks a concept as present that the annotator did not see, or the annotator saw a concept that the class-level label marks as absent.
Three sources contribute to this disagreement; Figure~\ref{fig:examples_two} shows examples, which occur frequently across the dataset.

\emph{Limited observability.} Many concepts that are typical of a species are not visible in individual images because of pose, occlusion, or resolution, yet the class-level label still marks them as present (Figure~\ref{fig:examples_two}, left: leg color and size).
Non-visibility rates are particularly high for fine-grained attributes such as wing or tail concepts, and these concepts consistently show lower agreement (Fig.~\ref{fig:cub_annotation_mismatch}(b) in Appendix~\ref{app:concept_ann_cub_koh}).

\emph{Aggregation.} Koh et al.~\cite{koh2020concept} construct class-level labels by majority voting: a concept is marked present for all images of a species if it is annotated as present in most of them, and concepts present in too few classes after voting are removed, which reduces the 312 attributes to 112. Majority voting removes intra-class variation, e.g., between sexes or plumages: a concept that the annotator saw in an image is marked absent if it occurs in only a minority of the species' images, and vice versa.

\emph{Annotation noise.} Each image is annotated by a single Amazon Mechanical Turk worker, so instance-level labels themselves contain errors: a concept can be marked as present although the corresponding part is not visible (Figure~\ref{fig:examples_two}, right: body and tail concepts annotated although only the head is visible).

These observations raise the question of what near-perfect concept accuracy under \textit{by-class} labels means. Concept predictors that follow the per-image evidence would agree with the class-level labels on only $\approx$82\% of pairs; accuracies well above this level are only reachable by predicting class-typical concepts that are not visible in the image, i.e., from cues to the species rather than from the concept itself.

\paragraph{Concept-Only Reference Accuracy}\label{par:concept_only_task_ceiling}
The annotation limits also bound how much task information the concepts can carry. We estimate this by training predictors directly on the binary ground-truth \textit{by-instance} concept vectors (\textit{by-class} vectors are excluded because they are identical per species and allow the lookup solution described above). Across linear logistic regression, a two-layer MLP, and gradient boosting (XGBoost), the best test accuracy is $37.5\%$ (two-layer MLP; Table~\ref{tab:concept_only_ceiling}, full results in Appendix~\ref{app:concept_ceiling}).
This value is not a strict upper bound: the instance-level annotations are noisy, and image models may recover graded visual evidence that the binary labels do not capture. We therefore treat $\approx$37\% as the centre of a \emph{reference band}: concept scores that carry only the information expressed by the annotated concepts should support task accuracy near it. Accuracies in the 70\%+ range therefore do not by themselves show that concept scores are bypassed or ungrounded, but they call for direct audits of where the additional information comes from (Section~\ref{sec:experiments}).

\begin{table}[t]
\centering
\caption{Concept-only reference accuracy on CUB. Classifiers receive only ground-truth instance-level binary concept annotations as input (all certainty levels, i.e., no concept is masked). We report the best test accuracy across concept-to-label predictors. The full table, including certainty-based masking, is available in Appendix~\ref{app:concept_ceiling}.}
\label{tab:concept_only_ceiling}
\begin{tabular}{lcc}
\toprule
Encoding & Best predictor & Test accuracy \\
\midrule
Binary ($\pm 1$) & MLP-2x256 & $37.50$ \\
\bottomrule
\end{tabular}
\end{table}

\section{How Task Information Can Enter Concept Scores}
\label{sec:cbm_architectural_problems}
Many CBMs report task accuracy well above the reference band of
Section~\ref{sec:ceilings}. Part of this gap may reflect legitimate
visual evidence that the binary annotations do not capture, but it
raises the question of which design choices let task information into
the concept scores without that information being backed by visual
evidence for the concepts. We identify four such design axes
(Section~\ref{sec:axes}). Each axis has a \emph{standard} setting, used
by common CBMs and CEMs, and a \emph{constrained} setting that closes
the corresponding pathway. Section~\ref{par:ablation} sweeps these axes
within a single model template (Section~\ref{sec:method}).

\subsection{Four Design Axes}
\label{sec:axes}

\textbf{Receptive field of the features.}
Many CUB concepts describe spatially localized attributes (e.\,g.,
\textit{has a brown head}, \textit{has an orange beak}), yet they are
standardly predicted from backbones such as ResNet18/50 whose receptive
fields cover most or all of the image. A concept predictor can then
exploit contextual or co-occurring evidence (e.\,g., background, pose,
or other body parts) instead of the attribute itself.
\emph{Constrained setting:} a BagNet backbone~\cite{brendel2019approximating}
with a fixed $k{\times}k$ receptive field per spatial location,
$k \in \{9, 17, 33\}$, so that every feature depends only on a local
image patch.

\textbf{Shared vs.\ concept-specific representations.}\label{par:shared_repr}
Standard CBMs predict all concepts from one shared embedding $g(x)$.
Since all concept predictors read the same features, the model is
encouraged to encode features that are jointly useful for many
concepts, which can lead to correlated activations: several concepts
fire together even when only a subset is present. This does not
directly imply leakage, but concept predictions are then not
disentangled and can share class-level information.
\emph{Constrained setting:} concept-specific heads, i.e., a separate
projection per concept on top of the backbone, so that each concept is
scored from its own feature vector rather than from an embedding shared
with all other concepts.

\textbf{Norm-dependent scoring.}
Standard concept scores are dot products between a shared embedding
and a concept weight vector,
\[
s_i(x) = w_i^\top g(x) = \|w_i\| \, \|g(x)\| \cos(\theta_i),
\]
so a score can increase through the feature norm $\|g(x)\|$ rather than
through better alignment $\cos(\theta_i)$ with the concept. On a
standard CBM, feature norm and the number of saturated concepts are
strongly correlated ($r=0.575$): large norms activate many concepts at
once. Normalizing both vectors removes saturation but collapses task
accuracy, indicating that magnitude carries class information
\textit{beyond} concept alignment (Appendix~\ref{app:norm}).
\emph{Constrained setting:} spherical scoring, i.e., cosine similarity
between $\ell_2$-normalized vectors. This constraint presupposes
concept-specific feature vectors. On the unit sphere, a single feature
vector can be close to two concept vectors only if these are close to
each other. A shared feature vector could therefore give high scores to
several distinct concepts at once only by making their concept vectors
nearly identical. Concept-specific vectors avoid this: each concept is
scored against its own feature vector, so several concepts can be
active in the same image region.

\textbf{Task supervision of the concept scores.}
In joint training, the label loss backpropagates into the concept
predictor. Concept scores can then be shaped toward class-discriminative
features regardless of whether these features support the concept;
Margeloiu et al.~\cite{margeloiu2021concept} report that jointly
trained CBMs encode target information beyond the concepts.
\emph{Constrained setting:} concept-only training, i.e., the first stage
of sequential training~\cite{koh2020concept}, with task information
measured afterwards from frozen concept scores.

\subsection{A Common Template for the Ablation}
\label{sec:method}
To vary these axes in a controlled way, we use a single concept-model
template; implementation details and an overview figure are in
Appendix~\ref{app:template}.
A backbone (ResNet50 or BagNet-$k$) computes a feature map
$g(x) \in \mathbb{R}^{C \times H \times W}$. A projection maps it to
concept latents $z_i(u,v) \in \mathbb{R}^d$, $i = 1, \dots, K$, at every spatial location
$(u,v)$, either through one head $\phi_i$ per concept or through a
single shared head. Each latent is compared with a learnable concept
vector $p_i \in \mathbb{R}^d$ and max-pooled into a concept score,
\[
s_i(x) = \max_{u,v} \, \hat{z}_i(u,v)^\top \hat{p}_i \in [-1,1],
\qquad \hat{z}_i = z_i / \|z_i\|, \;\; \hat{p}_i = p_i / \|p_i\|,
\]
i.e., spherical scoring. Concept-only variants minimize a weighted binary cross-entropy
\begin{equation}
- \sum_{(x_j,c_j) \in D} \sum_{i=1}^{K}
\lambda_{j,i}
 \left[
c_{j,i}\, \log\bigl(\hat{c}_{j,i}\bigr)
+ \bigl(1 - c_{j,i}\bigr)\, \log\bigl(1-\hat{c}_{j,i}\bigr)
\right],
\label{EQ:Weighted-BEC-Concepts}
\end{equation}
where $\hat{c}_{j,i} = \sigma(s_i(x_j))$. With \textit{by-class} labels,
$\lambda_{j,i} = 1$. With \textit{by-instance} labels, $\lambda_{j,i}$
down-weights uncertain annotations according to the annotator certainty
$\kappa_{j,i} \in \{1,2,3,4\}$ (Section~\ref{sec:prelim}), with
$\lambda = 0.1, 0.4, 0.75, 1$ respectively. Joint variants add a
cross-entropy loss on a linear class head.

\textbf{Measuring class information in concept scores.}
For every variant, task accuracy is measured by a bias-free linear
probe $\hat{y} = W^\top s$ fit on frozen concept scores. The probe is
the only path from the image to a class prediction, so no variant can
bypass its concept scores; the probe accuracy therefore measures how
much class information the scores themselves carry. We omit the bias so
that the probe has no class preference when no concept evidence is
present.

We refer to the configuration in which every axis is set to its
constrained value---BagNet-33, concept-specific heads, spherical
scoring, and concept-only training---as the \emph{fully constrained
configuration}, denoted \modelname{} in tables. It is not proposed as a
competitive model but is the end point of the sweep; the choice of
BagNet-33 among the local backbones is itself an outcome of the
receptive-field sweep (Section~\ref{par:ablation}).

\section{Experiments}
\label{sec:experiments}
The experiments follow the two failure modes of
Section~\ref{sec:intro}. Leakage probes first test whether CBM and CEM predictions bypass their exposed concept scores
(Section~\ref{par:leakage_probe}). A controlled ablation over the design
axes of Section~\ref{sec:cbm_architectural_problems} then measures how
much class information enters the concept scores themselves
(Section~\ref{par:ablation}). Finally, spatial localization and part
occlusion test whether concept scores are driven by the image regions
that should support them (Sections~\ref{sec:spatial_faithfulness_sec}
and~\ref{sec:occlusion_faithfulness}).

\textbf{Setup.} We audit Joint CBM~\cite{koh2020concept} and
CEM~\cite{espinosa2022concept}, both with a ResNet18 backbone on full
images (Appendix~\ref{app:training}), and the template variants of
Section~\ref{sec:method}, which use a bounding-box crop. Results are mean $\pm$ std over three seeds
with early stopping on validation concept accuracy. All experiments use
CUB-200-2011~\citep{wah2011caltech} with the $K=112$ concepts of Koh et
al.~\cite{koh2020concept} and 200 classes, under the \textit{by-class}
and \textit{by-instance} regimes of Section~\ref{sec:prelim}. Both
regimes use the same 112 concepts: \textit{by-instance} labels are the
original per-image CUB annotations of these concepts, so the two regimes
differ only in the labels, not in the concept set. Chance level is
$0.5\%$ (1/200 classes).

\subsection{Do CBM and CEM Predictions Bypass the Concept Scores?}\label{par:leakage_probe}
We begin with the first failure mode: whether a model's native predictions can be reconstructed from its exposed concept scores. Let $s(x)$ denote the continuous concept scores predicted by a model, and let $d(x)$ denote their discretized version. We fit linear probes
$\pi_{\mathrm{cont}}: s(x)\mapsto y$ and $\pi_{\mathrm{disc}}: d(x)\mapsto y$
using only concept outputs. For CEM, we also fit a probe $\pi_{\mathrm{emb}}: e(x)\mapsto y$ on the internal concept embedding $e(x)$. $\mathrm{Acc}_{\mathrm{probe,disc}}$ and $\mathrm{Acc}_{\mathrm{probe,cont}}$ quantify the ease of reconstructing class labels from discrete and continuous predicted concepts, respectively. $\mathrm{Acc}_{\mathrm{native}}$ denotes the model's own task-head accuracy. All probes are fit with \texttt{scikit-learn} logistic regression to convergence (no solver warnings across settings), so probe gaps are not an artifact of optimizer underfitting.

We report three gaps: the \emph{bypass gap} $\Delta_{\mathrm{bypass}}
= \mathrm{Acc}_{\mathrm{native}} - \mathrm{Acc}_{\mathrm{probe,cont}}$,
capturing whether the model relies on information outside the predicted
concepts (a positive value indicates a non-concept decision path); the
\emph{encoding gap} $\Delta_{\mathrm{enc}} = \mathrm{Acc}_{\mathrm{probe,cont}}
- \mathrm{Acc}_{\mathrm{probe,disc}}$, measuring information lost when
concepts are discretized; and, for CEM, the \emph{embedding leakage}
$\Delta_{\mathrm{emb}} = \mathrm{Acc}_{\mathrm{probe,emb}} -
\mathrm{Acc}_{\mathrm{probe,cont}}$, quantifying label information
present in internal representations but not in the exposed concept
outputs.

\begin{table*}[t]
\centering
\small
\setlength{\tabcolsep}{5pt}
\caption{Concept accuracy, AUC, and leakage probe results
(\% accuracy, mean $\pm$ std). Nat.: native task accuracy;
Cont./Disc./Emb.: linear probes on continuous, discretized, and
internal-embedding concept outputs. $\Delta_b$, $\Delta_{enc}$,
$\Delta_{emb}$ quantify bypass, encoding, and embedding leakage.
$\dagger$: fully constrained configuration of the ablation
(Section~\ref{sec:method}), shown for reference; its predictions come
only from the probe, so $\Delta_b=0$ by construction. Gaps are computed before rounding.}
\label{tab:leakage_probe}
\resizebox{\textwidth}{!}{
\begin{tabular}{l|cc|ccccccc}
\toprule
Model
& Concept Acc.
& AUC
& Nat.
& Cont.
& Disc.
& Emb.
& $\Delta_b$
& $\Delta_{enc}$
& $\Delta_{emb}$ \\
\midrule
\multicolumn{10}{c}{\textit{by-instance}} \\
\midrule
CEM
& $79.83$\scriptsize$\pm0.43$
& $66.57$\scriptsize$\pm0.19$
& $67.52$\scriptsize$\pm2.48$
& $52.21$\scriptsize$\pm1.90$
& $38.76$\scriptsize$\pm2.15$
& $68.56$\scriptsize$\pm1.23$
& $15.31$
& $13.45$
& $16.34$ \\
CBM
& $80.22$\scriptsize$\pm0.13$
& $63.97$\scriptsize$\pm0.48$
& $58.20$\scriptsize$\pm2.67$
& $57.16$\scriptsize$\pm1.48$
& $52.06$\scriptsize$\pm2.35$
& --
& $1.04$
& $5.11$
& -- \\
\modelname{}$^\dagger$
& $74.84$\scriptsize$\pm0.07$
& $70.20$\scriptsize$\pm0.13$
& --
& $25.23$\scriptsize$\pm0.48$
& $21.84$\scriptsize$\pm1.45$
& --
& $0.00$
& $3.39$
& -- \\
\midrule
\multicolumn{10
}{c}{\textit{by-class}} \\
\midrule
CEM
& $94.89$\scriptsize$\pm0.15$
& $90.23$\scriptsize$\pm0.54$
& $76.91$\scriptsize$\pm0.18$
& $69.68$\scriptsize$\pm0.83$
& $69.54$\scriptsize$\pm1.07$
& $74.98$\scriptsize$\pm0.74$
& $7.23$
& $0.14$
& $5.30$ \\
CBM
& $94.78$\scriptsize$\pm0.12$
& $89.39$\scriptsize$\pm0.08$
& $72.64$\scriptsize$\pm0.16$
& $69.03$\scriptsize$\pm0.68$
& $68.45$\scriptsize$\pm0.47$
& --
& $3.61$
& $0.58$
& -- \\
\modelname{}$^\dagger$
& $90.35$\scriptsize$\pm0.38$
& $83.17$\scriptsize$\pm0.06$
& --
& $42.82$\scriptsize$\pm0.13$
& $45.97$\scriptsize$\pm1.12$
& --
& $0.00$
& $-3.15$
& -- \\
\bottomrule
\end{tabular}
}
\end{table*}

Table~\ref{tab:leakage_probe} reveals qualitatively different behaviors across models.
Across both evaluation settings, CEM shows a substantial bypass gap (e.g., $+7.23$ \textit{by-class} and $+15.31$ \textit{by-instance}), indicating that its native predictions cannot be reconstructed from exposed concepts alone. Moreover, the embedding probe consistently outperforms the concept probe (e.g., $\Delta_{\mathrm{emb}}=+5.30$ \textit{by-class} and $+16.34$ \textit{by-instance}), demonstrating that class information is stored in internal representations that are not reflected in the concept outputs. This gap widens significantly in the more challenging \textit{by-instance} setting, suggesting increased reliance on non-concept information.

The embedding probe also serves as a capacity check on the probing procedure itself: fit on the full internal representation, $\pi_{\mathrm{emb}}$ matches (\textit{by-instance}: $68.56\%$ vs.\ native $67.52\%$) or nearly matches (\textit{by-class}: $74.98\%$ vs.\ native $76.91\%$, within $2$pp) native accuracy. Since this is the same linear probing procedure applied to a strictly higher-dimensional representation, its ability to recover native-level accuracy indicates that the probe is not the limiting factor; the smaller gaps observed for the continuous and discretized concept-score probes are therefore unlikely to reflect probe underfitting rather than genuine information loss at the exposed interface.

CBM has a smaller bypass gap than CEM (e.g., $+3.61$ \textit{by-class}), but does not eliminate it, indicating residual dependence on non-concept pathways. We note that this gap is not directly comparable to CEM's: CBM's label predictor $f$ is trained jointly with the backbone rather than fit post-hoc on frozen scores, so part of the gap may reflect this training-regime mismatch rather than a distinct non-concept pathway; we do not have an analogous embedding-level probe for CBM to disentangle the two. Additionally, non-trivial encoding gaps (e.g., $+0.58$ \textit{by-class} and $+5.11$ \textit{by-instance}) suggest that discretized concepts lose task-relevant information, particularly in the \textit{by-instance} regime.

Table~\ref{tab:leakage_probe} also lists the fully constrained
configuration of Section~\ref{sec:method} for reference. Its class
predictions come only from the probe, so $\Delta_{\mathrm{bypass}}=0$
holds by construction; its probe accuracy is discussed with the
ablation below.

\subsection{Where Does the Class Information in Concept Scores Come From?}\label{par:ablation}
The probes address the first failure mode. The second one does not
require a bypass: the concept scores can themselves carry class
information that is not backed by visual evidence for the concepts they represent. We
therefore sweep the design axes of
Section~\ref{sec:cbm_architectural_problems} within the template of
Section~\ref{sec:method} and measure, for each configuration, concept
quality and the class information that a probe recovers from frozen
concept scores (Table~\ref{tab:ablations}). Because no configuration
can bypass its scores, differences in probe accuracy reflect
differences in what the scores encode, not in how the scores are used.

\begin{table}[t]
    \centering
    \small
    \setlength{\tabcolsep}{3.5pt}
    \caption{Systematic ablation of the design axes of
Section~\ref{sec:cbm_architectural_problems} (mean $\pm$ std,
\textit{by-class} annotations). All configurations use the template of
Section~\ref{sec:method} with spherical scoring;
task accuracy is measured by linear probes on continuous (Cont.) and
discretized (Disc.) frozen concept scores, so no configuration can
bypass its scores. $\dagger$: fully constrained configuration (\modelname{}).}
    \label{tab:ablations}
    \begin{tabular}{lll|cc|cc}
    \toprule
     & & & \multicolumn{2}{c|}{\textit{Concept quality}}
     & \multicolumn{2}{c}{\textit{Task Acc (probe)}} \\
    \textbf{Backbone} & \textbf{Heads} & \textbf{Training}
     & Conc.\ Acc & Conc.\ AUC
     & Cont. & Disc. \\
    \midrule
    \multicolumn{7}{l}{\textit{Head sharing}} \\
    ResNet50  & Shared & Concept-only
        & $82.53$\scriptsize$\pm0.16$ & $84.03$\scriptsize$\pm0.49$
        & $64.66$\scriptsize$\pm0.62$ & $60.57$\scriptsize$\pm0.57$ \\
    ResNet50  & Specific & Concept-only
        & $95.79$\scriptsize$\pm0.13$ & $92.85$\scriptsize$\pm0.21$
        & $72.11$\scriptsize$\pm0.55$ & $72.59$\scriptsize$\pm0.68$ \\
    BagNet-33 & Shared & Concept-only
        & $77.21$\scriptsize$\pm0.52$ & $70.47$\scriptsize$\pm3.79$
        & $20.63$\scriptsize$\pm5.30$ & $13.40$\scriptsize$\pm4.33$ \\
    \midrule
    \multicolumn{7}{l}{\textit{Task supervision}} \\
    BagNet-33 & Specific & Joint
        & $70.58$\scriptsize$\pm3.16$ & $72.43$\scriptsize$\pm2.40$
        & $54.95$\scriptsize$\pm4.29$ & $55.29$\scriptsize$\pm3.60$ \\
    \midrule
    \multicolumn{7}{l}{\textit{Receptive field (specific heads, concept-only)}} \\
    BagNet-33$^\dagger$ & Specific & Concept-only
        & $90.35$\scriptsize$\pm0.38$ & $83.17$\scriptsize$\pm0.06$
        & $42.82$\scriptsize$\pm0.13$ & $45.97$\scriptsize$\pm1.12$ \\
    BagNet-17 & Specific & Concept-only
        & $84.03$\scriptsize$\pm0.61$ & $79.95$\scriptsize$\pm0.27$
        & $32.04$\scriptsize$\pm0.70$ & $31.81$\scriptsize$\pm0.54$ \\
    BagNet-9 & Specific & Concept-only
        & $76.21$\scriptsize$\pm0.47$ & $73.96$\scriptsize$\pm0.28$
        & $15.92$\scriptsize$\pm1.23$ & $12.62$\scriptsize$\pm1.27$ \\
    \bottomrule
    \end{tabular}
\end{table}

\textbf{Receptive field is the main lever.}
With concept-specific heads and concept-only training fixed, probe
accuracy falls monotonically as the receptive field shrinks: $72\%$ for
ResNet50, $43\%$ for BagNet-33, $32\%$ for BagNet-17, and $16\%$ for
BagNet-9. Most of the class information that ResNet50 concept scores
carry therefore comes from visual context beyond a local patch. The
sweep also shows that locality has a cost of its own: concept accuracy
drops from $95.8\%$ (ResNet50) to $90.4\%$ (BagNet-33), $84.0\%$
(BagNet-17), and $76.2\%$ (BagNet-9), because very small patches miss
the spatial extent needed for shape, pattern, or multi-part color
attributes. BagNet-33 is the most local setting in the sweep that still
predicts concepts reliably, which is why the fully constrained
configuration uses it.

\textbf{Concept-specific heads are a precondition, not a cost.}
Replacing concept-specific heads with a single shared head lowers both
concept quality and probe accuracy, with ResNet50 ($95.8 \to 82.5\%$
concept accuracy; $72 \to 65\%$ probe accuracy) and much more strongly
with BagNet-33 ($90.4 \to 77.2\%$; $43 \to 21\%$). The shared-head
variants keep spherical scoring, so a shared latent at a given location
can align only with concept vectors that are close to each other
(Section~\ref{sec:cbm_architectural_problems}). Distinct concepts that
occur in the same patch, such as the color and pattern of a wing, then
compete for one direction. This interaction is strongest when each
location sees only a small patch, which is consistent with the larger
drop for BagNet-33. Concept-specific heads are therefore needed for
usable concept scores under spherical scoring rather than being a
source of lost accuracy.

\textbf{Norm-dependent scoring.}
All template variants use spherical scoring, so the ablation does not
isolate the scoring axis. Evidence for it comes from the normalization
study on a standard CBM (Appendix~\ref{app:norm}), where removing the
norm dependence removes concept saturation but also collapses task
accuracy, i.e., feature magnitude carries class information beyond
concept alignment.

\textbf{Task supervision injects class information at the expense of
the concepts.}
Adding joint task training to BagNet-33 with concept-specific heads
raises probe accuracy from $43\%$ to $55\%$ while lowering concept
accuracy from $90.4\%$ to $70.6\%$. Even with local features and
separate heads, task gradients push the concept scores toward
class-discriminative features that the concept labels do not support.

\textbf{The constrained end point.}
Setting every axis to its constrained value yields the fully
constrained configuration (\modelname{}). Its probe accuracy
($43\%$ \textit{by-class}, $25\%$ \textit{by-instance};
Table~\ref{tab:leakage_probe}) lies within the reference band around
$37\%$: once the pathways of Section~\ref{sec:cbm_architectural_problems}
are closed, the concept scores carry roughly as much class information
as the ground-truth concept annotations do. Its lower task accuracy is
therefore not a failure of this configuration but the expected outcome
of the sweep.

The grid is not fully factorial, and swapping the backbone may change
more than the receptive field. We therefore read the sweep as
mechanistic evidence for where class information enters the concept
scores, not as an exact causal decomposition of CBM or CEM accuracy.

\begin{figure*}[t]
    \centering

    \begin{minipage}{0.49\linewidth}
        \centering
        \includegraphics[width=\linewidth]{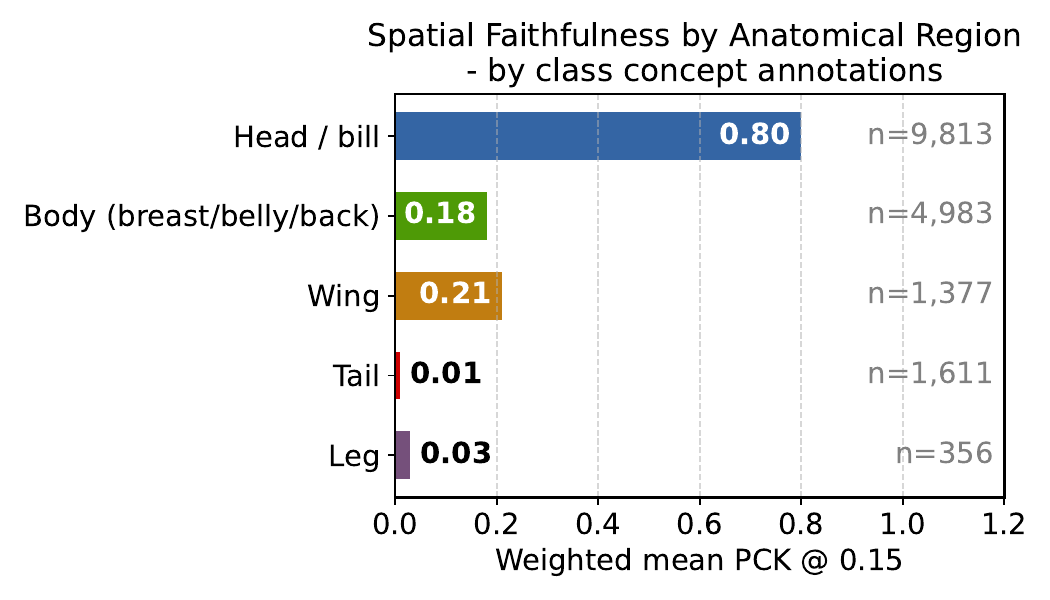}
    \end{minipage}
    \hfill
    \begin{minipage}{0.49\linewidth}
        \centering
        \includegraphics[width=\linewidth]{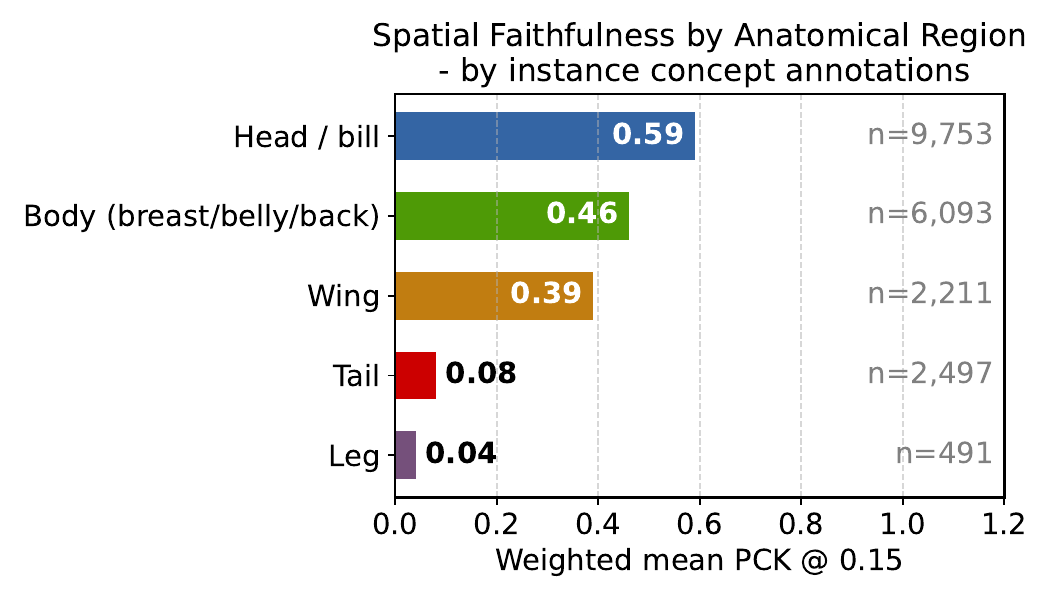}
    \end{minipage}

    \caption{
     Spatial faithfulness (PCK@0.15) by anatomical region. Models trained with \textit{by-class} annotations (left) exhibit strong head bias, while \textit{by-instance} supervision (right) yields more balanced localization across body parts.
    }
    \label{fig:anatomy_spatial_faith}
\end{figure*}

\subsection{Is Locality Enough for Grounding? Spatial Localization}
\label{sec:spatial_faithfulness_sec}
Closing the pathways of Section~\ref{sec:cbm_architectural_problems}
limits how much class information the concept scores carry, but it does
not guarantee that each score is driven by the evidence a human would
use to recognize the concept. Local features, however, make this
question measurable: because every BagNet-33 feature depends only on a
$33{\times}33$ patch, a concept score can be traced back exactly to the
patch that produced it, without attribution approximations. We use this
to test whether the concept activations of the fully constrained
configuration peak near the expected anatomical part. Training provides
no incentive for this: part keypoints are never used, and the only
signal is the image-level concept label applied to the max-pooled score.
Correct localization therefore has to emerge from which patches best
predict the concept label, which is exactly what the test probes. CBM and CEM are
excluded because their global features do not expose comparable
spatial concept maps. We compare the configuration trained with
class-level and with instance-level annotations, isolating the effect
of the supervision signal. We localize
each concept at its peak activation and, following Huang et
al.~\cite{huang2024concept}, compare it to the CUB part keypoints,
reporting PCK@0.15.\footnote{Percentage of Correct Keypoints---PCK@$\alpha$: The fraction of predicted concept activation locations that fall within a threshold distance $\alpha$ (normalized by the bird bounding-box diagonal) of the ground-truth CUB keypoint for that concept.} Evaluation is restricted to samples
where the concept is present, annotator certainty is above threshold,
the model predicts the concept, and the keypoint is visible (full
protocol in Appendix~\ref{app:spatial_faith}).
Figure~\ref{fig:anatomy_spatial_faith} shows a clear difference between
training regimes. Under instance-level supervision, localization is
more balanced across parts (PCK@0.15: $0.59$ head, $0.46$ body, $0.08$
tail, $0.04$ leg), whereas class-level supervision produces a strong
head bias (PCK@0.15: $0.80$ head, $0.18$ body). The same holds for a
stricter test that asks whether an activation lies closer to the mapped
part than to a competing part sharing the same attribute: $0.569$ under
instance-level supervision versus near-chance $0.519$ under class-level
supervision.

\textit{Head-proxy behavior.} Under class-level supervision, body,
wing, and tail concepts are predicted predominantly from the head
region (head/body PCK ratio $4.4\times$). Since all images of a species
share the same concept vector, the head, which is highly discriminative
for species identity, becomes a proxy for every concept of that
species. Instance-level supervision weakens this shortcut, reducing the
ratio to $1.3\times$. Locality therefore constrains \emph{where} the
evidence for a concept can come from, but the supervision signal
decides \emph{which} local evidence is used.

\subsection{Do Concept Scores Respond to Their Evidence? Part Occlusion}\label{sec:occlusion_faithfulness}
Spatial localization only applies to models with local features. As a
behavioral test that applies to any model, we adapt the part-removal
tests of~\cite{huang2024concept,jeon2025locality}: we blur the
anatomical part mapped to each positive concept and compare the
resulting change of the concept score against random and
unrelated-part masks. A grounded
concept score should drop when its supporting part is removed, and more
so than under control masks. We compare the fully constrained
configuration with CEM.

\begin{table}[t]
\centering
\small
\setlength{\tabcolsep}{5pt}
\caption{Part-occlusion audit under class-level CUB supervision. Scores are macro-averages over mapped anatomical groups. Target drop is the change in the exposed concept score when the mapped part is blurred; target--random and target--unrelated are specificity contrasts against control masks. Flip rate reports changes in the native class prediction after target occlusion. Backbone, preprocessing, score scale ($[-1,1]$ for \modelname{}, $[0,1]$ for CEM), and evaluated pairs differ between rows, so contrasts should be read within each row.}
\label{tab:occlusion_results}
\begin{tabular}{lrrrrr}
\toprule
Model & Pairs & Target drop & Target--random & Target--unrelated & Flip rate \\
\midrule
\modelname{} & 69,228 & 0.437 & 0.355 & 0.153 & 29.1 \\
CEM & 25,994 & 0.034 & 0.017 & 0.003 & 19.4 \\
\bottomrule
\end{tabular}
\end{table}

Because the two models differ in backbone, preprocessing, score scale,
and evaluated pairs (Section~\ref{sec:experiments}), we compare
within-model contrasts rather than raw score drops across models
(Table~\ref{tab:occlusion_results}). For the fully constrained
configuration, whose scores are cosine similarities in $[-1,1]$,
occluding the mapped part lowers the concept score by $0.437$ on
average, well above the drop under random masks (target--random
contrast $0.355$). For CEM, whose scores are probabilities in $[0,1]$,
occluding the mapped part lowers the concept score by only $0.034$,
barely more than a random mask ($0.017$ contrast), even though
occlusion flips the native class prediction in $19.4\%$ of evaluated
pairs. CEM's predictions thus change under visual intervention while
its exposed concept scores barely reflect the removed evidence---precisely the
situation in which a human inspecting or correcting these scores would
be misled.

The occlusion results also corroborate the head-proxy behavior. Under
class-level supervision, the fully constrained configuration is highly
sensitive to head masks (target-score drop $0.755$; flip rate
$49.4\%$) but much less sensitive to masks of non-head parts (target
drop $0.135$; flip rate $13.9\%$): its concept scores do respond to
local evidence, but that evidence is often the head rather than the
annotated part.

\section{Discussion and Implications}
\label{sec:discussion}
Our main conclusion is not that high CUB accuracy is meaningless, nor
that concept models should ignore task performance. Rather, CUB task
accuracy is underdetermined as evidence for concept-based reasoning:
the same number can arise from visible concept scores, internal
embeddings, non-local image features, shared representations, or
task-gradient shaping. For CBMs, this distinction matters because the
model's scientific and practical promise is not only classification,
but an inspectable and intervenable concept interface. In line with
the view that some leakage can be benign when the concept set is
incomplete~\cite{espinosa2026defense}, we do not argue that every
source of additional accuracy should be removed; we argue that it
should be identified before the accuracy is credited to the concepts.

The concept-only reference provides the first point of calibration. The
CUB concept annotations explain far less task performance than modern
CBM/CEM variants achieve. Because the annotations are noisy and
incomplete, this is not a hard ceiling, but it is a warning that task
accuracy alone cannot tell us whether a model reasons through the
concepts a user can inspect.

The probes and the ablation then locate the additional accuracy. CEM's
native predictions are not fully reconstructable from its scalar
concept scores, and its internal embeddings contain the missing class
information---the first failure mode. The ablation shows that the
second failure mode does not need a bypass: within a single template,
the class information carried by the concept scores rises from the
level of the annotations to CEM-like values when the receptive field
is enlarged, and rises again under joint task training, which at the
same time lowers concept accuracy. High accuracy on CUB can therefore
be recovered by relaxing these constraints, but doing so changes what
the concept scores mean. The receptive-field sweep also shows that
locality is not an unqualified virtue: very small receptive fields miss
the visual extent needed for shape, pattern, or multi-part color
attributes, while global ones make it easy to predict a concept from
correlated evidence elsewhere in the image. The useful design question
is therefore not ``the smallest possible receptive field,'' but whether
the evidence used for a concept matches the evidence a user expects
that concept to use.

The grounding results show that closing these pathways is necessary but
not sufficient. Under class-level supervision, local concept scores
often predict non-head attributes from the head, which is highly
discriminative for species identity on CUB. Such head-proxy behavior is
exactly the kind of mismatch that accuracy numbers hide: a concept score
can be predictive at the scalar level while its evidence comes from a
different region than a human would expect, so a user who corrects it
after inspecting the annotated part is not correcting what the model
used. Instance-level supervision reduces this bias but does not remove
all localization failures, especially for small or peripheral parts
such as tails and legs. Part occlusion makes the same point for CEM:
its predictions change under targeted masks while its exposed concept
scores remain nearly flat, so inspecting those scores would not reveal
why the prediction changed.

\textbf{Implications for evaluating CBMs.}
Our results suggest reporting, alongside native task accuracy, (i) the
accuracy of a probe on frozen concept scores and the resulting bypass
gap, (ii) where applicable, the class information in internal concept
embeddings, (iii) the concept-only accuracy of the annotations used for
supervision as a reference band, and (iv) for models with spatial
concept maps, localization or occlusion checks against part
annotations. Together, these quantities separate accuracy that is
carried by grounded concept scores from accuracy that arises elsewhere.

\textbf{The role of concept annotations.}
Our findings highlight the central role of concept annotations in
determining what a faithful CBM can learn and explain. On CUB, the
available concepts are noisy, partially observable, and incomplete, and
class-level labels disagree with the per-image annotations often enough
that near-perfect concept accuracy under these labels itself signals
prediction from species cues rather than from visible concepts.
This is not a failure of concept bottlenecks in general. It means that
concept vocabularies and annotation protocols define the interpretive
contract a model can satisfy. More precise and visually grounded
annotations could support higher faithful accuracy, while unsupervised
or generative concept discovery methods~\cite{melis2018sen,
NEURIPS2019_adf7ee2d, li2018deep, marconato2022glancenets,
yang2023language, oikarinen2023label} may expand the concept interface;
their faithfulness, however, should be audited with the same care.

\textbf{Limitations.}
We focus on CUB-200-2011 with 112 processed concepts, so the results
should not be read as a universal estimate of leakage or faithful
accuracy for all CBM datasets. The concept-only reference depends on noisy
single-annotator instance labels encoded as binary concept vectors; it is
a reference band, not a ground-truth ceiling. The ablation grid is not
fully factorial, and swapping backbones may change more than the
receptive field, so it supports mechanistic rather than exact causal
conclusions. The CBM and CEM baselines use a different backbone
(ResNet18) and preprocessing (no bounding-box crop) than the template
configurations, and their concept scores live on different scales;
we therefore interpret occlusion effects within each model rather than
comparing raw effects across models. The spatial evaluation is limited to concepts with
well-defined anatomical anchors and to correctly predicted concept
instances, and therefore does not capture false positives or all forms
of mislocalized evidence. Finally, we argue that ungrounded concept
scores undermine interventions but do not evaluate interventions
directly; our grounding analyses test their precondition.

\bibliographystyle{plain}
\bibliography{biblio}

\newpage
\appendix
\section{Analysis of Concept Annotation Reliability on CUB}
\label{app:concept_ann_cub_koh}

This appendix provides additional analysis supporting the results in Sec .~\ref{sec:ceilings}, focusing on the reliability of concept annotations under the standard CUB preprocessing pipeline~\cite{koh2020concept}.

\paragraph{Data and preprocessing}
We use the processed CUB dataset introduced by Koh et al.~\cite{koh2020concept}, with 112 binary concepts derived from the original 312 attributes via filtering, grouping, binarization, and majority voting.
For each image, we extract:
(i) class-level concept annotations used in prior work,
(ii) instance-level concept annotations from the original dataset~\cite{wah2011caltech}, and
(iii) annotator certainty scores provided in~\cite{wah2011caltech}.

\paragraph{Agreement metric}
Let $y_c \in \{0,1\}$ denote the class-level label for concept $c$, and let
$\hat{y}_{i,c} \in \{0,1\}$ denote the instance-level annotation for concept $c$ in image $i$.
Each instance-level annotation is associated with a certainty score $\kappa_{i,c} \in \{1,2,3,4\}$.

We evaluate agreement as a function of a certainty threshold $t$, retaining only annotations with $\kappa_{i,c} \geq t$. We define the set of retained image--concept pairs as
\[
K_t = \{(i,c) \mid \kappa_{i,c} \geq t\}.
\]

Agreement is then defined as
\[
\mathrm{Agreement}(t)
=
P\!\left(\hat{y}_{i,c} = y_c \;\middle|\; (i,c) \in K_t\right)
=
\frac{1}{|K_t|}
\sum_{(i,c)\in K_t}
\mathbf{1}\!\left[\hat{y}_{i,c}=y_c\right].
\]

This corresponds to the fraction of retained image--concept pairs for which instance-level annotations match class-level labels. It defines an oracle ceiling for class-level evaluation: a perfect model that predicts the true image-level concepts would achieve exactly this accuracy when evaluated against class-level labels.

\begin{figure*}[t]
    \centering

    \begin{minipage}{0.48\linewidth}
        \centering
        \includegraphics[width=\linewidth]{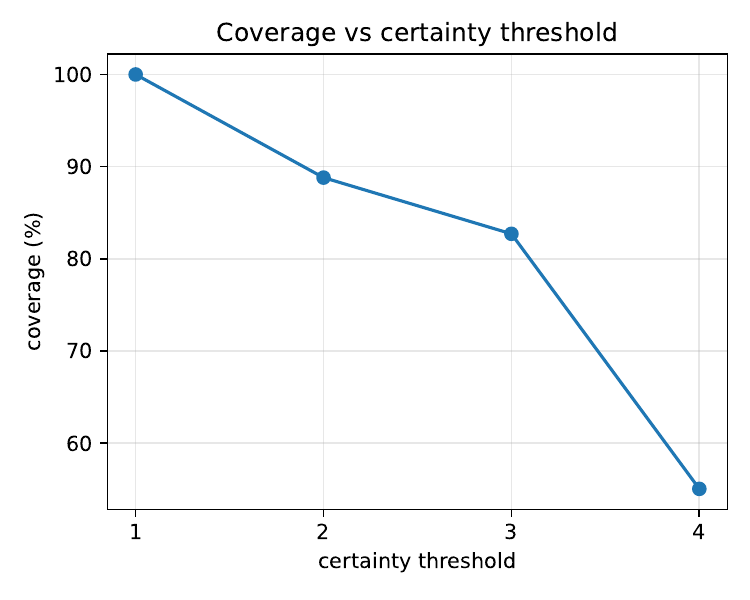}
        \subcaption{Coverage vs. certainty}
    \end{minipage}
    \hfill
    \begin{minipage}{0.48\linewidth}
        \centering
        \includegraphics[width=\linewidth]{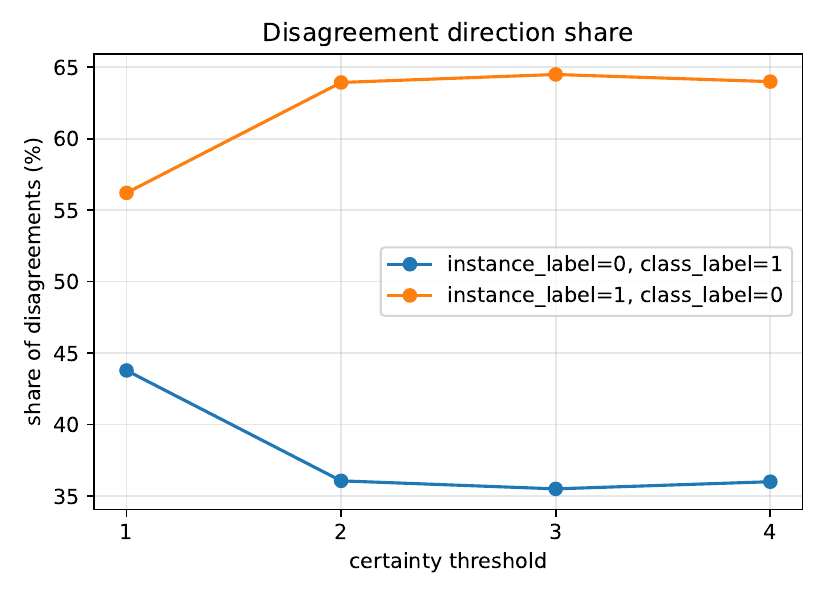}
        \subcaption{Disagreement direction}
    \end{minipage}

    \caption{
    Additional analysis of the class--instance annotation mismatch on CUB.
    (a) Increasing the certainty threshold substantially reduces coverage, dropping to roughly $55\%$ at the highest threshold.
    (b) Disagreement direction reveals two sources of mismatch: class-positive concepts that are not visible in individual images, and instance-positive concepts suppressed by class-level aggregation.
    }
    \label{fig:cub_appendix_mismatch}
\end{figure*}

\begin{figure}[t]
    \centering

    \begin{minipage}{0.45\linewidth}
        \centering
        \includegraphics[width=\linewidth]{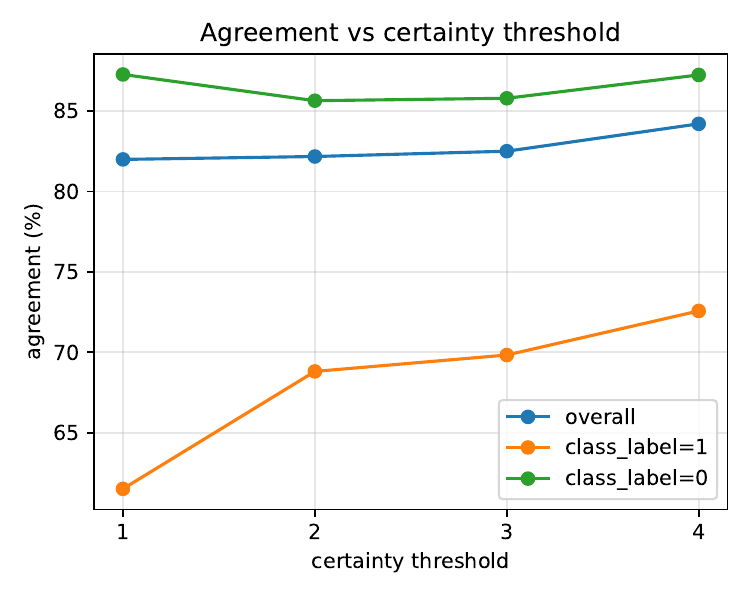}
        \subcaption{Agreement vs. certainty}
    \end{minipage}
    \hfill
    \begin{minipage}{0.50\linewidth}
        \centering
        \includegraphics[width=\linewidth]{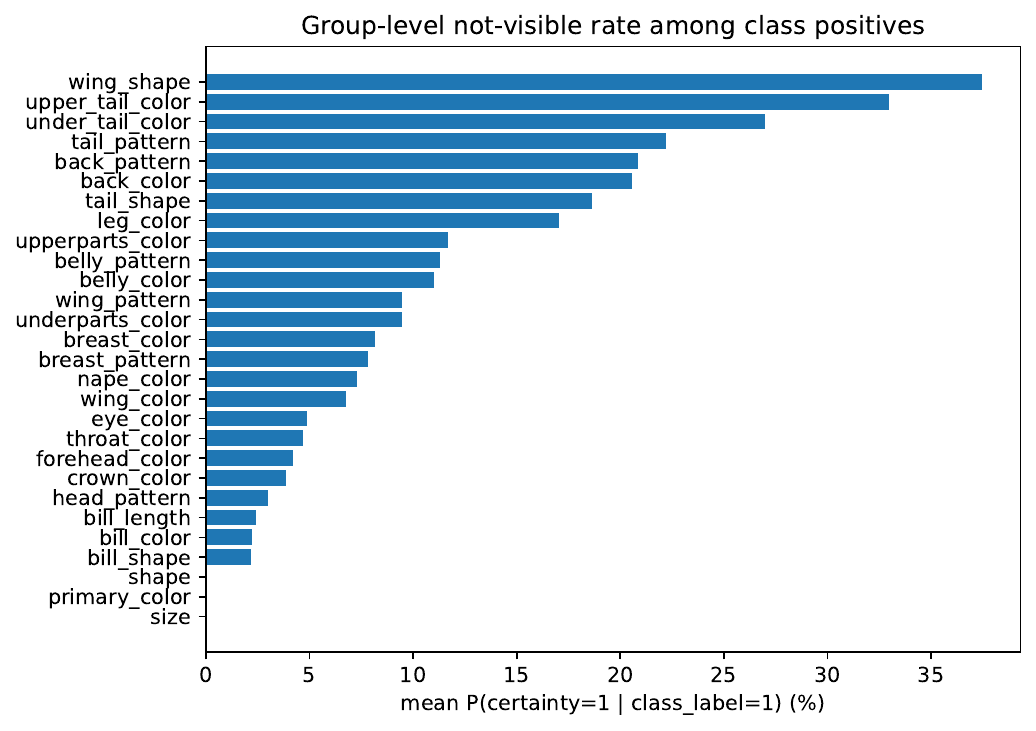}
        \subcaption{Non-visibility by concept group}
    \end{minipage}

    \caption{
    Annotation mismatch in CUB. 
    (a) Agreement between instance-level and class-level concept labels remains bounded around $82$--$84\%$ across certainty thresholds, showing that filtering uncertain annotations only marginally improves alignment. \textit{class\_label=0} denotes disagreement between class-level (green) and instance-level annotations, while \textit{class\_label=1} denotes agreement; results over both cases are shown in blue.
    (b) Non-visibility rates vary substantially across concept groups, especially for fine-grained attributes, indicating that limited observability is a major source of mismatch.
    }
    \label{fig:cub_annotation_mismatch}
\end{figure}

\paragraph{Coverage--agreement trade-off}
We define Coverage as the fraction of image--concept pairs that survive the certainty filter $\kappa \geq t$. 
More formally: $\text{Coverage}(t) = \frac{|K_t|}{|K_1|}$ 
At $t = 1$, coverage is $100\%$ by definition (no pairs are rejected). At $t = 4$, only pairs for which the annotator was fully certain are retained.
Figures ~\ref{fig:cub_annotation_mismatch}(a) and Fig.~\ref{fig:cub_appendix_mismatch}(a) show the trend of Agreement and Coverage with respect to the certainty threshold $t$.
Specifically, we can see that increasing $t$ leads to only modest improvements in Agreement (from $\approx 82\%$ to $\approx 84\%$), while substantially reducing Coverage (from $\approx 100\%$ to $\approx 50\%$). 
This confirms that the mismatch between class- and instance-level annotations is structural rather than driven by low-confidence labels.

\paragraph{Disagreement decomposition}
To better understand the sources of annotation errors, we analyze the direction of Disagreement between image-level and class-level labels. 
Let $D_t$ denote the set of disagreeing image--concept pairs in $K_t$.

We distinguish two complementary types of disagreement. 
The quantity $\delta_{0 \to 1}$ (class-positive miss) measures the fraction of disagreements where the image-level annotation indicates absence, while the class-level label indicates presence. 
This form of disagreement typically occurs when a concept characteristic of the class is not visible in a particular image (e.g., due to pose, occlusion, or viewpoint).

Conversely, $\delta_{1 \to 0}$ (class-negative contradiction) measures the fraction of disagreements where the image-level annotation indicates presence, but the class-level label indicates absence. 
This reflects when class-level concept prevalence is thresholded (e.g., at $50\%$), forcing minority instances within a class to be incorrectly labeled as negative.

Formally, these quantities are defined as:
\begin{equation}
\delta_{0 \to 1}(t) = \mathbb{P}\big(\hat{y}_{i,c} = 0,\; y_c = 1 \;\mid\; (i,c) \in D_t\big)
\end{equation}

\begin{equation}
\delta_{1 \to 0}(t) = \mathbb{P}\big(\hat{y}_{i,c} = 1,\; y_c = 0 \;\mid\; (i,c) \in D_t\big)
\end{equation}

By construction, $\delta_{0 \to 1}(t) + \delta_{1 \to 0}(t) = 1$.

Figure~\ref{fig:cub_appendix_mismatch}(b) shows that the Disagreements are clearly asymmetric and become more so with higher certainty.
$\delta_{1 \to 0}$ dominates (~64–65\%), meaning most conflicts occur when the image shows the concept but the class label says it is absent.

\paragraph{Concept observability}
We next analyze how agreement varies across concept groups. A \emph{concept group} is a set of semantically related attributes describing a common part or property of the object (e.g., \textit{wing\_shape}, \textit{bill\_color}), which are analyzed jointly due to their shared observability and noise characteristics.

We find that concepts that are difficult to observe (e.g., fine-grained shape or tail attributes) exhibit substantially lower agreement than consistently visible attributes such as global color or size (Figure~\ref{fig:cub_annotation_mismatch}(b)).
This pattern closely mirrors the variation in non-visibility rates across groups, indicating that observability is a primary driver of annotation noise.

\paragraph{Summary}
Overall, we find that:
(i) Agreement between instance- and class-level annotations is fundamentally bounded ($\approx 82\%$),
(ii) Increasing annotation certainty yields only marginal improvements,
(iii) Disagreement arises from both observability limitations and aggregation artifacts, and
(iv) Annotation reliability varies substantially across concepts.

These results support the conclusion that the concept representation used in standard CUB benchmarks is inherently noisy and imposes a structural limit on concept-based learning performance.

\section{Norm Effects in Concept Scoring}
\label{app:norm}
\begin{table}[t]
\setlength{\tabcolsep}{4pt}
\centering
\small
\caption{Controlled normalization study on CUB. 
We independently normalize $g$ and $w$ and report score decomposition, norm-saturation correlations, and downstream performance. In the standard configuration, $\sim$99.7\% of score variance is explained by the norm product. Fixing one norm shifts the dependency to the other, while fixing both removes norm-driven variability. 
}
\begin{tabular}{lcccc}
\toprule
 & Unconstrained & $g$ normalized & $w$ normalized & Both normalized \\

 \midrule
\multicolumn{1}{l}{} & \multicolumn{4}{c}{\textbf{Concept score decomposition}} \\  

Norm product contribution
& 0.997 & 0.651 & 0.995 & 0.000 \\

Cosine contribution 
& 0.003 & 0.349 & 0.005 & 1.000 \\

\midrule

\multicolumn{1}{l}{} & \multicolumn{4}{c}{\textbf{Norm--saturation correlation}} \\  

Feature norm vs. saturation
& 0.575 & $-0.017$ & 0.520 & <0.001 \\

Weight norm vs. saturation
& 0.197 & 0.382 & $-0.039$ & <0.001 \\

\midrule
\multicolumn{1}{l}{} & \multicolumn{4}{c}{\textbf{Downstream performance}} \\  

Concept accuracy 
& 0.828 & 0.812 & 0.836 & 0.816 \\

Task accuracy 
& 0.398 & 0.183 & 0.404 & 0.034 \\

\bottomrule
\end{tabular}
\label{tab:norm_ablation}
\end{table}
In this appendix, we analyze the role of feature and weight magnitudes in concept-bottleneck models (CBMs) and provide additional evidence supporting the claims in Section~\ref{sec:cbm_architectural_problems}.

\paragraph{Score decomposition.}
Concept-bottleneck models (CBMs)~\cite{koh2020concept} are composed as$f \circ g \circ h : \mathcal{X} \xrightarrow{h} \mathcal{Z} \xrightarrow{g} \mathcal{C} \xrightarrow{f} \mathcal{Y}$, with predictions made from concept outputs rather than directly from image features. 
In standard CBMs, concepts are predicted via linear classifiers on the shared representation $z = h(x)$. The score for concept $i$ is $s_i(x) = w_i^\top z$ where $w_i$ is the concept weight vector. 
This admits the decomposition\begin{equation}s_i(x) = \|z\| \, \|w_i\| \cos(\theta_i),\end{equation}where $\theta_i$ is the angle between $z$ and $w_i$, separating semantic alignment ($\cos(\theta_i)$) from the scaling effects of the feature norm $\|z\|$ and the weight norm $\|w_i\|$.

Concept scores in standard CBMs take the form
\begin{equation}
s_i(x) = w_i^\top g(x) = \|g(x)\| \, \|w_i\| \cos(\theta_i),
\end{equation}
where $g(x)$ is the shared feature embedding, $w_i$ is the concept weight vector, and $\theta_i$ is the angle between $g(x)$ and $w_i$.
This decomposition separates semantic alignment ($\cos(\theta_i)$) from two independent scaling factors: the feature norm $\|g(x)\|$ and the weight norm $\|w_i\|$.

\paragraph{Experimental setup.}
To isolate the contribution of each factor, we perform a controlled normalization study in which we independently normalize $\|g(x)\|$ and $\|w_i\|$ to $1$. 
This results in four configurations: (i) standard CBM, (ii) g normalized ($\|g(x)\| = 1$), (iii) w normalized ($\|w_i\| = 1$), and (iv) both normalized (pure cosine similarity).
All models are trained under identical conditions on CUB dataset by ~\cite{koh2020concept}.

\begin{figure*}[t]
\centering
\begin{minipage}{0.48\textwidth}
    \centering
    \includegraphics[width=\linewidth]{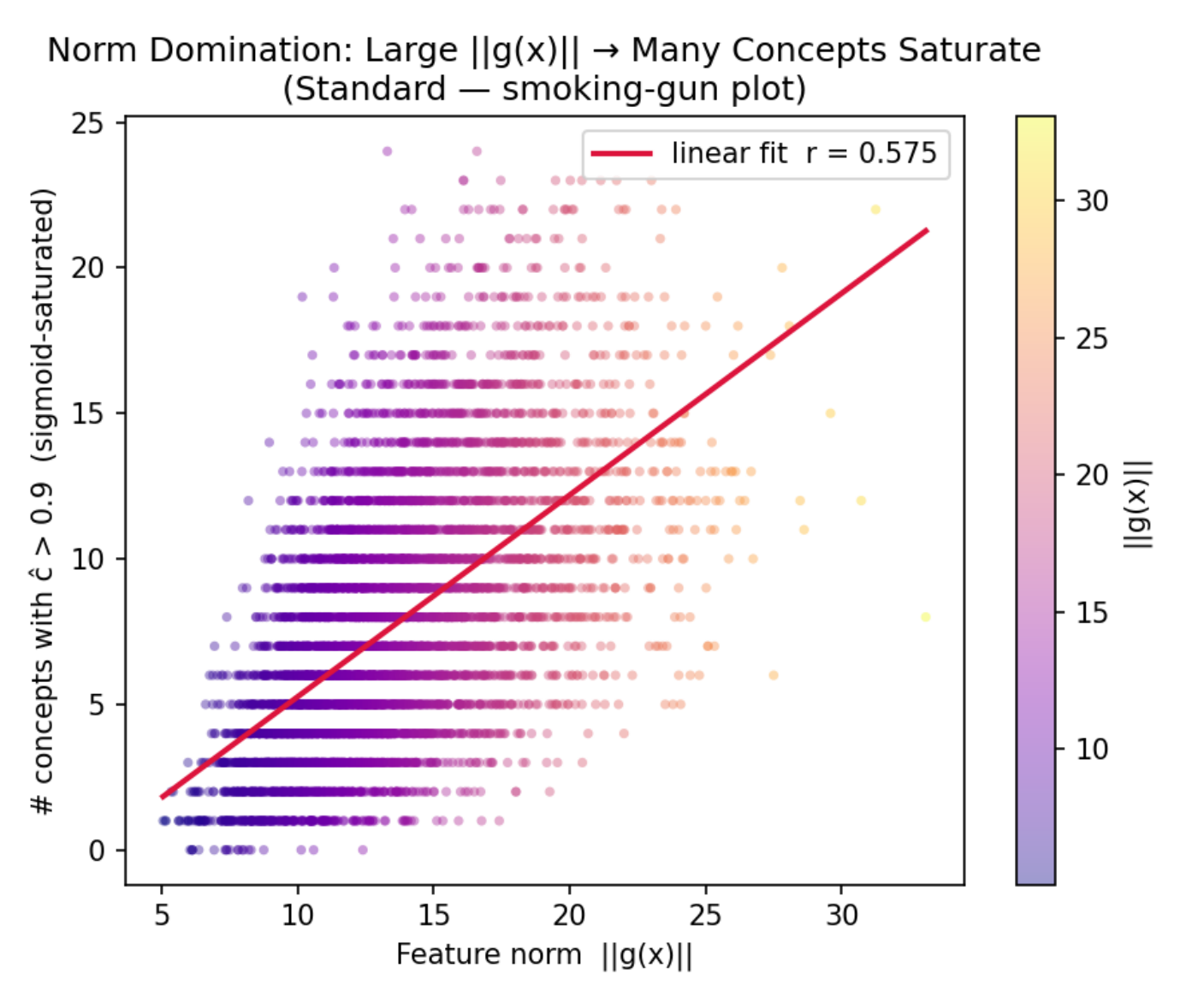}
\end{minipage}
\hfill
\begin{minipage}{0.48\textwidth}
    \centering
    \includegraphics[width=\linewidth]{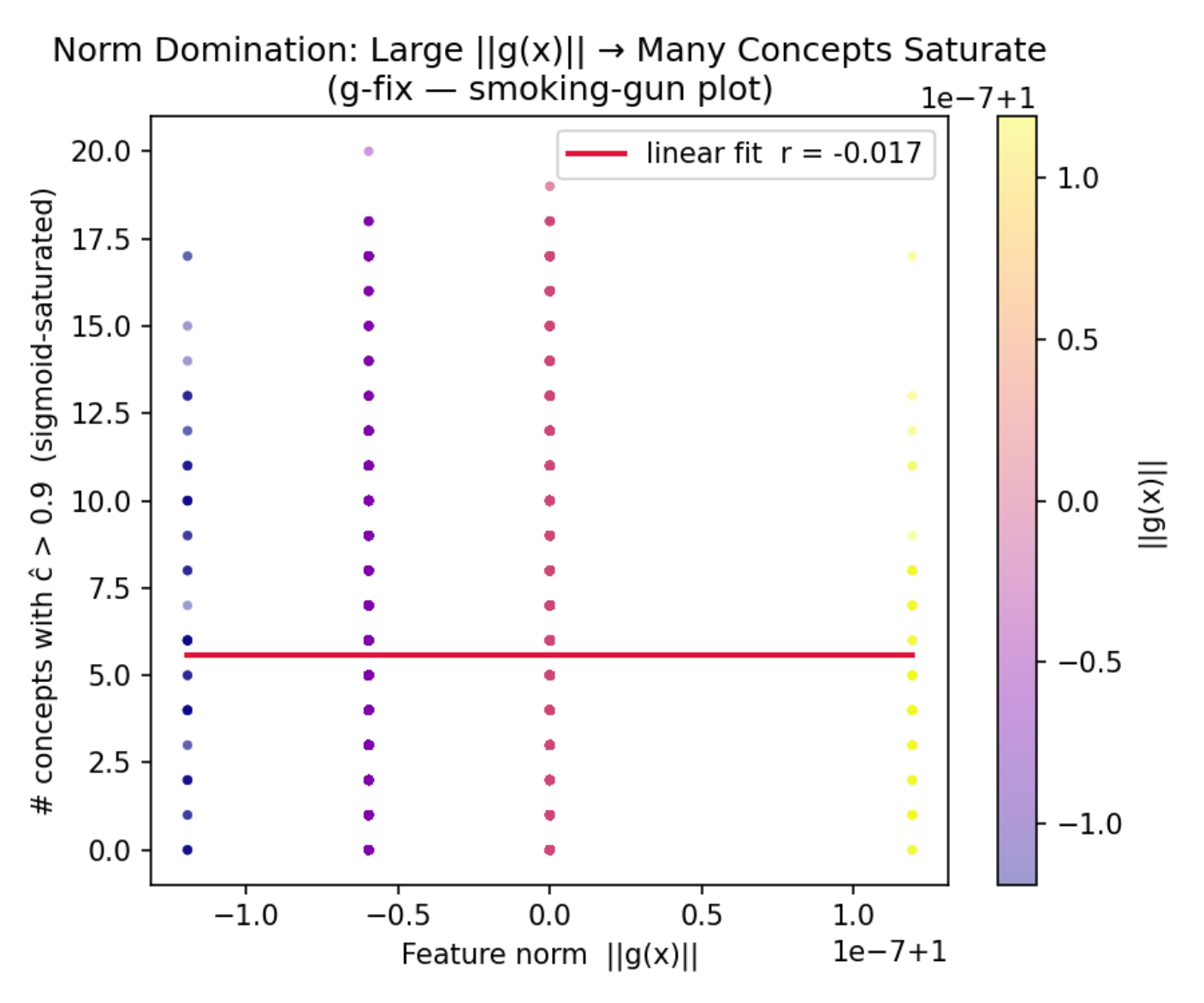}
\end{minipage}

\vspace{0.5em}

\begin{minipage}{0.48\textwidth}
    \centering
    \includegraphics[width=\linewidth]{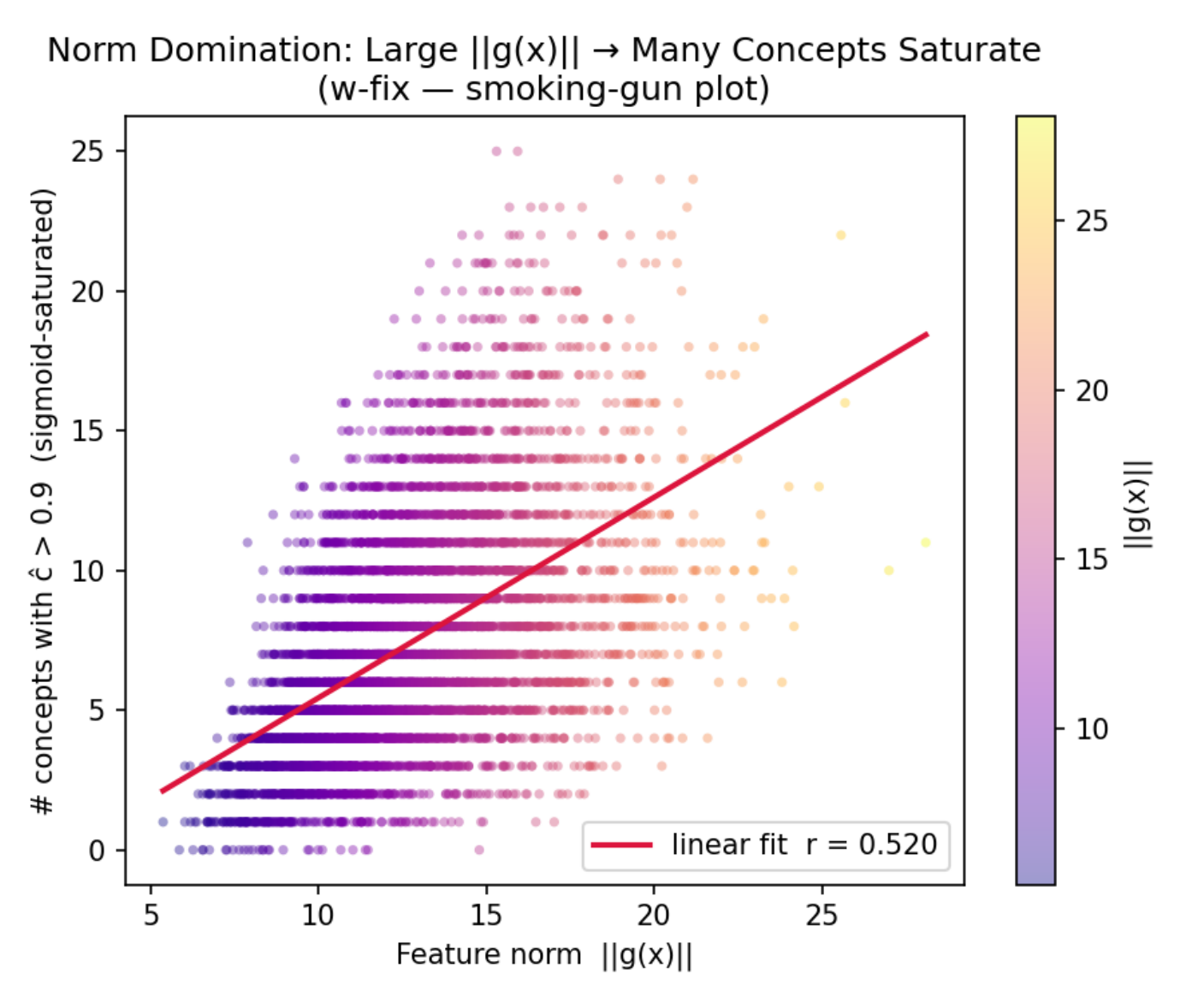}
\end{minipage}
\hfill
\begin{minipage}{0.48\textwidth}
    \centering
    \includegraphics[width=\linewidth]{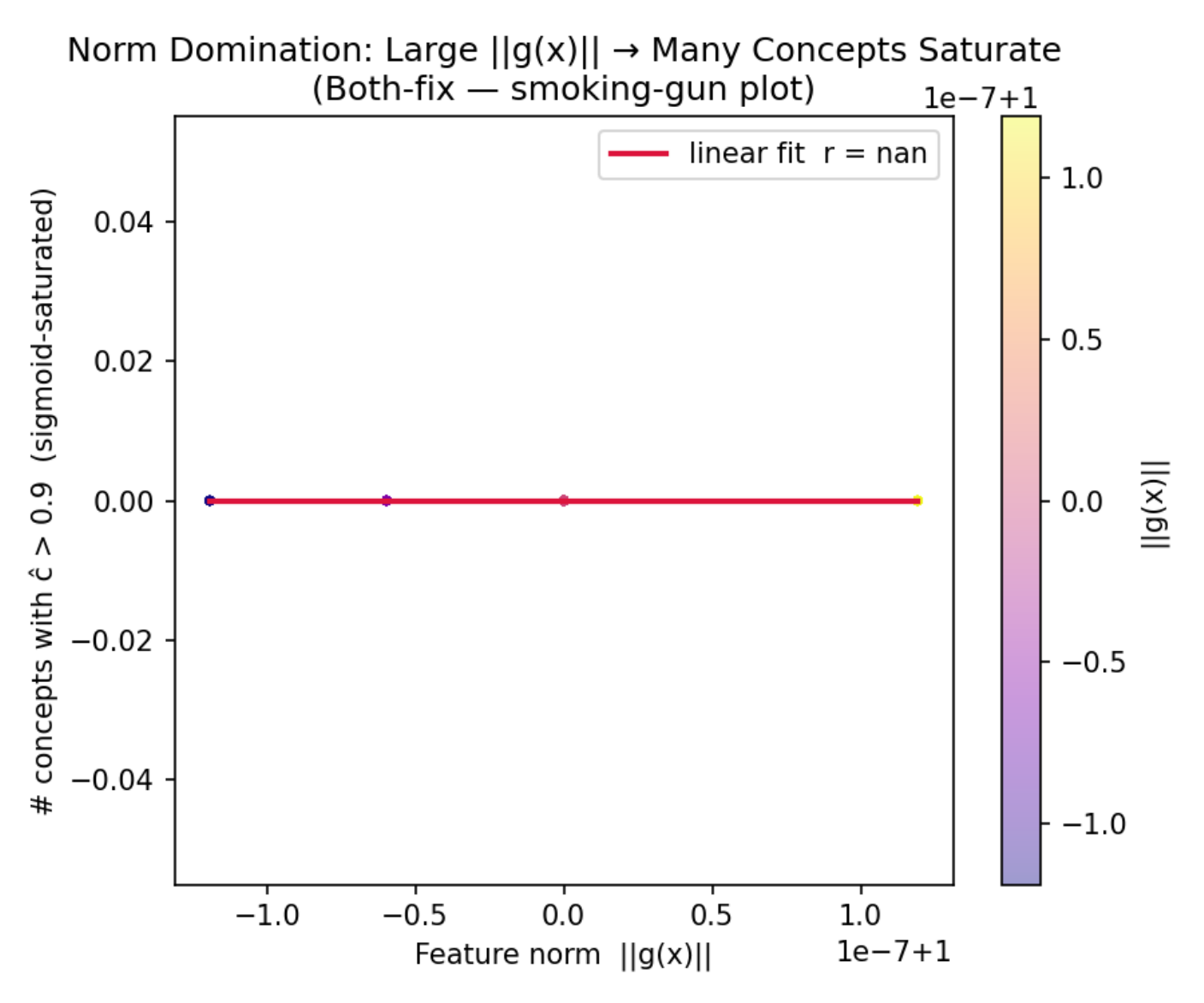}
\end{minipage}

\caption{
Feature norm $\|g(x)\|$ versus the number of saturated concepts across the four ablation settings.
\textit{Top-left:} standard CBM.
\textit{Top-right:} g-fix, where $\|g(x)\|=1$.
\textit{Bottom-left:} w-fix, where $\|w_i\|=1$.
\textit{Bottom-right:} both-fix, where both norms are constrained.
The standard model exhibits a strong positive relation between feature norm and saturation.
Fixing $\|g(x)\|$ removes this dependency, while fixing only $\|w_i\|$ leaves it largely intact.
When both norms are fixed, saturation disappears entirely.
}
\label{fig:norm_saturation}
\end{figure*}

\paragraph{Norm-driven saturation.}
Figure~\ref{fig:norm_saturation} presents the relationship between feature norm $\|g(x)\|$ and the number of saturated concepts for all four configurations.
The standard model (top-left) exhibits a strong positive correlation ($r = 0.575$), indicating that larger feature norms systematically lead to more saturated predictions.

Fixing $\|g(x)\|$ (top-right) removes this dependency entirely ($r \approx 0$), confirming that feature magnitude is a primary driver of saturation.
In contrast, fixing $\|w_i\|$ (bottom-left) leaves the dependency largely unchanged ($r = 0.520$), showing that constraining weights alone is insufficient.
When both norms are fixed (bottom-right), saturation disappears completely, as scores are bounded and no longer depend on magnitude.

\paragraph{Compensatory behavior.}
Constraining a single norm does not eliminate magnitude effects, as the model compensates through the remaining degree of freedom.
When $\|g(x)\|$ is fixed, weight norms increase substantially (mean $0.978 \rightarrow 4.299$), and their correlation with saturation rises ($r = 0.197 \rightarrow 0.382$).
Conversely, fixing $\|w_i\|$ removes variability in weight norms but leaves feature norms free, preserving their influence on saturation.
This behavior, visible across the four panels, indicates that the two norm factors act as interchangeable channels for controlling score magnitude.

\paragraph{Variance decomposition.}
We quantify the relative contribution of magnitude and alignment by decomposing score variance.
In the standard model, the norm product $\|g(x)\| \cdot \|w_i\|$ explains $99.7\%$ of the variance, while cosine similarity contributes only $0.3\%$.
Fixing $\|g(x)\|$ increases the cosine contribution to $34.9\%$, but magnitude still dominates, whereas fixing $\|w_i\|$ has negligible effect.
Only when both norms are fixed does cosine similarity account for the full variance ($100\%$), confirming that standard CBMs rely almost entirely on magnitude rather than alignment.

\paragraph{Impact on performance.}
While removing norm effects leads to a geometrically well-defined formulation, it comes at a high cost.
Task accuracy drops from $0.398$ in the standard model to $0.034$ under both-fix, indicating that magnitude is critical for class separation in high-dimensional feature space.
Partial fixes exhibit intermediate behavior: w-fix preserves performance, while g-fix degrades it due to compensatory growth in weight norms.

\paragraph{Summary.}
These results support three key conclusions:
(i) Concept scores in standard CBMs are dominated by magnitude rather than semantic alignment,
(ii) Feature and weight norms contribute independently and compensate for each other when constrained, and
(iii) Fully removing magnitude eliminates saturation but severely degrades performance, motivating softer forms of norm control.

\section{Concept-Only Reference Accuracy: Full Results}
\label{app:concept_ceiling}
In this appendix, we provide additional details and robustness analyses for the
concept-only reference accuracy reported in Section~\ref{par:concept_only_task_ceiling}.

\paragraph{Experimental protocol.}
To estimate the task performance achievable from concepts alone, we derive a dataset of pairs $(c^*,y)$ from the CUB dataset by Koh et al.~\cite{koh2020concept}, where $c^*$ is the masked-sign concept vector constructed from the ground-truth instance-level annotations of image $x$, and $y$ denotes the corresponding class label~\cite{wah2011caltech}.
Specifically, concepts with annotator certainty below a threshold $t$ are masked to zero; retained concepts take values in $\{-1, +1\}$ according to the binary annotation.
We vary the certainty threshold $t \in \{1,2,3,4\}$, producing multiple
annotation regimes with different degrees of visual reliability. At $t=1$
no concept is masked, so the encoding is the plain binary annotation
recoded from $\{0,1\}$ to $\{-1,+1\}$; all three predictor families are
invariant to this affine recoding. The reference accuracy of
Section~\ref{par:concept_only_task_ceiling} is obtained at $t=1$.

\paragraph{Model families.}
To ensure that the estimated reference accuracy is not limited by the capacity of the
concept-to-label mapping, we evaluate multiple predictor classes:
(i) Linear logistic regression,
(ii) A two-layer MLP with $256$ hidden units per layer, and
(iii) Gradient-boosted decision trees (XGBoost).
All models are selected based on validation performance and evaluated on the
held-out test set.

\begin{table}[t]
\centering
\caption{Concept-only task accuracy on CUB using ground-truth concepts.
We report test accuracy (\%) for different concept encodings, certainty thresholds $t$,
and concept-to-label predictors. The best result per encoding is highlighted in bold.}
\label{tab:concept_only_full}
\begin{tabular}{llcccc}
\toprule
Encoding & Model & $t=1$ & $t=2$ & $t=3$ & $t=4$ \\
\midrule

\multirow{3}{*}{Masked sign}
& Linear & 32.46 & 32.46 & 31.50 & 19.62 \\
& MLP (2x256) & \textbf{37.50} & \textbf{37.50} & 35.45 & 25.34 \\
& XGBoost & 36.02 & 36.02 & 34.05 & 22.82 \\

\bottomrule
\end{tabular}

\end{table}

\paragraph{Results.}
Table~\ref{tab:concept_only_full} reports the full results across all
configurations. 
We make the following three observations:
\textbf{(i) Low concept-only accuracy.}
Across all encodings and predictors, the maximum achievable test accuracy
remains below $40\%$, with a peak of $37.5\%$. This demonstrates that the
annotated instance-level concepts alone do not explain the classification
performance reported by many CBMs, motivating direct audits of what
additional information the trained models use.
\textbf{(ii) Robustness across model classes.}
The reference accuracy is stable across linear, neural, and tree-based predictors.
While more expressive models (MLP, XGBoost) slightly outperform linear models,
the improvement is marginal, indicating that the limitation arises from the
concept representation itself rather than the classifier capacity.
\textbf{(iii) Effect of annotation filtering.}
Performance varies with the certainty threshold $t$. Moderate filtering
($t=1,2$) yields the highest accuracy, while aggressive filtering ($t=4$)
significantly degrades performance. 
This reflects a trade-off between annotation noise and information loss: removing uncertain concepts improves reliability but reduces the amount of available signal.

\paragraph{Implications.}
These results establish the center of the concept-only reference band for concept-based classification on CUB.
Because the band has fuzzy boundaries (single annotator per image, subjectivity of certainty scores), it should be treated as a reference point rather than a strict bound. Models operating substantially above it may use legitimate graded visual evidence, non-local proxy cues, or internal information not exposed in scalar concept scores; distinguishing these cases is the purpose of the leakage probes and ablations in Section~\ref{sec:experiments}.

\section{Spatial Faithfulness}
\label{app:spatial_faith}
For visual concepts, interface faithfulness also has a spatial component: a model may correctly predict a concept while relying on evidence from an unexpected region. To evaluate this, we measure \emph{spatial faithfulness}, i.e., whether concept activations localize near the anatomical region associated with the concept.

\paragraph{Architecture and backprojection.}
The fully constrained configuration (\modelname{}) uses a BagNet-33 backbone, producing a spatial feature map $f(x) \in \mathbb{R}^{U \times V \times C}$ with stride $s=8$ and receptive field size $r=33$. For location $(u,v)$, the receptive field center in input coordinates is:
\[
c_x = \frac{r - 1}{2} + s \cdot v, 
\qquad
c_y = \frac{r - 1}{2} + s \cdot u.
\]
For each concept $c$, we localize it via the peak activation:
\[
(u^*, v^*) = \arg\max_{u,v} f_c(x)_{u,v},
\]
and map it to $(c_x, c_y)$.

\paragraph{Setup.}
We compare \modelname{} trained with class-level (Koh) annotations and instance-level annotations. Concepts are mapped to CUB keypoints across 17 anatomical regions (see mapping below). For bilateral parts (e.g., wings), the nearest visible keypoint is used.

\paragraph{Evaluation filtering.}
A pair $(x,c)$ is evaluated only if: (i) the concept is present, (ii) annotator certainty $\geq 3$, (iii) the model predicts it as present, and (iv) the keypoint is visible. This yields $\approx$68k (Koh) and $\approx$77k (instance-level) valid pairs.

\paragraph{Coordinate normalization.}
Distances are computed in $224 \times 224$ model-input space and normalized by the bounding-box diagonal:
\[
d_{\text{norm}} = \frac{\| (c_x, c_y) - (x_{\text{kp}}, y_{\text{kp}}) \|_2}{224}.
\]

\paragraph{PCK (Percentage of Correct Keypoints).}
We measure localization using
\[
\text{PCK}(\alpha) = 
\frac{1}{N}
\sum_{i=1}^N 
\mathbf{1}\big[ d_{\text{norm}}^{(i)} < \alpha \big].
\]
We evaluate $\alpha \in [0.05, 0.50]$ and report $\mathrm{PCK}@0.15$. At $224 \times 224$, $\alpha=0.15$ corresponds to $\approx$34 pixels (about four BagNet strides), balancing robustness and spatial precision.

\paragraph{Spatial discrimination.}
To assess fine-grained localization, we define a relative metric. For concept pairs $c_1, c_2$ sharing a descriptor but mapped to different parts:
\[
\text{Disc}(c_1, c_2) = 
\mathbb{E}\left[
\mathbf{1}\big(
d(\text{peak}_{c_1}, \text{part}_{c_1}) 
<
d(\text{peak}_{c_1}, \text{part}_{c_2})
\big)
\right].
\]
A score of $0.5$ is chance. Values $>0.5$ indicate that activations are closer to the mapped region even for visually similar concepts. This is stricter than PCK, as it requires discriminating between nearby parts.

\begin{table}[t]
\centering
\small
\caption{PCK across thresholds $\alpha$ for \modelname{} trained with class-level and instance-level annotations. Instance-level supervision consistently improves localization, particularly at stricter thresholds.}
\begin{tabular}{l|c|c}
\toprule
$\alpha$ & \modelname{} (class-level) & \modelname{} (instance-level) \\
\midrule
0.05 & 0.113 & 0.130 \\
0.10 & 0.284 & 0.326 \\
0.15 & 0.422 & 0.476 \\
0.20 & 0.541 & 0.596 \\
0.30 & 0.748 & 0.777 \\
0.35 & 0.829 & 0.841 \\
0.40 & 0.883 & 0.886 \\
0.45 & 0.919 & 0.921 \\
0.50 & 0.946 & 0.946 \\
\bottomrule
\end{tabular}

\label{tab:appendix_pck_curve}
\end{table}

\paragraph{Results.}
Table~\ref{tab:appendix_pck_curve} shows that both models achieve $\mathrm{PCK}@0.50 \approx 0.95$, indicating on-object activations. However, fine localization differs: $\mathrm{PCK}@0.15$ improves from $0.42$ (class-level) to $0.48$ (instance-level).
Table~\ref{tab:appendix_concept_to_cub} also shows the extracted anatomical groups. Refer to Appendix Section~\ref{app:concept_cub_mapping} for more details.

\begin{table}[h]
\centering
\small
\caption{PCK@0.15 by anatomical region. Instance-level supervision reduces head bias and improves localization on body parts.}
\begin{tabular}{lccc}
\toprule
Region & Koh & Instance & $\Delta$ \\
\midrule
Head / Bill & 0.80 & 0.59 & $-0.21$ \\
Body        & 0.18 & 0.46 & $+0.28$ \\
Wing        & 0.21 & 0.39 & $+0.18$ \\
Tail        & 0.01 & 0.08 & $+0.07$ \\
Leg         & 0.03 & 0.04 & $+0.01$ \\
\bottomrule
\end{tabular}
\label{tab:appendix_anatomy_pck}
\end{table}

\paragraph{Anatomical breakdown.}
Table~\ref{tab:appendix_anatomy_pck} highlights a strong anatomical bias under class-level supervision: activations concentrate on head regions ($0.80$) while remaining low on body parts ($0.18$), indicating reliance on class-correlated features. Instance-level supervision substantially reduces this bias, improving localization on body and peripheral regions and yielding a more balanced spatial distribution. Instance-level supervision reduces this bias (body $0.46$), lowering the head/body ratio from $4.4\times$ to $1.3\times$.
Tail and leg concepts remain challenging (PCK@0.15 $<0.09$), likely due to small spatial extent, peripheral location, and grid resolution.

\paragraph{Spatial discrimination results.}
The class-level model operates near chance (0.519), while instance-level supervision achieves 0.569, confirming improved discrimination between nearby anatomical regions.

\paragraph{Interpretation.}
These results show that high concept accuracy does not imply spatially faithful reasoning. Class-level supervision encourages shortcuts via global correlations (e.g., head features), whereas instance-level supervision encourages activations closer to the mapped anatomical regions.

Importantly, architectural locality alone is insufficient: BagNet constrains \emph{where information comes from}, but not \emph{where the model attends}. Spatial faithfulness therefore depends critically on the supervision signal.

\paragraph{Limitations.}
Evaluation is conditioned on correct predictions and does not capture false positives. Localization is also limited by the discrete grid (stride $=8$). Finally, performance varies across regions due to dataset properties (occlusion, size, discriminativeness).

\section{Codebook Fidelity}\label{app:exp_codebook}
In this section, we evaluate whether model predictions can be explained as
decisions over discrete concept representations. While the probe analysis in
Section~\ref{par:leakage_probe} tests whether predictions are reconstructible from
continuous concept scores, we now assess whether decisions are consistent with
a \emph{codebook-based reasoning} mechanism operating on discretized concepts.

\paragraph{Method.}
We construct a class-level concept codebook $\{k_c\}_{c=1}^{C}$, where each
class $c$ is associated with a binary concept vector derived from the
class-level CUB annotations. For each test image $x$, we first discretize the
predicted concept vector $d(x)$ and binarize it via thresholding
($d_i(x) > 0$). We then assign a predicted class by selecting the nearest
codebook entry under Hamming distance:
\[
\hat{y}_{\mathrm{code}}(x)
=
\arg\min_{c} \; \mathrm{Ham}(d(x), k_c).
\]

\paragraph{Metrics.}
We report three diagnostics:
\begin{itemize}
    \item \textbf{Mean Hamming distance:} average distance between $d(x)$ and
    the nearest class code.
    \item \textbf{Nearest-code / native agreement:} fraction of test samples
    where the nearest codebook class matches the model's native prediction.
    \item \textbf{Nearest-code / true-class match:} classification accuracy
    obtained via nearest-code assignment.
\end{itemize}

\begin{table}[t]
\centering
\caption{Codebook-fidelity diagnostics.}
\label{tab:codebook}
\begin{tabular}{lcc}
\toprule
Metric & \modelname{} & CEM \\
\midrule
Mean Hamming distance & $5.35$ & $3.93$ \\
Nearest-code / native agreement & $0.03$ & $79.50$ \\
Nearest-code / true-class match & $45.05$ & $70.50$ \\
\bottomrule
\end{tabular}
\end{table}

\paragraph{Results.}
Table~\ref{tab:codebook} reports the codebook-fidelity diagnostics.
For the fully constrained configuration, the nearest-code classification accuracy is $45.05\%$, closely
matching both the continuous-concept probe accuracy ($44.99\%$) and the
discrete-concept probe accuracy ($47.58\%$) on the checkpoint analyzed
here (seed-averaged values: Table~\ref{tab:ablations}). This consistency indicates that predictions can be well approximated as nearest-code decisions in the discrete
concept space, supporting the interpretation that the model routes decisions
through the exposed concept representation.

The nearest-code / native agreement is close to zero for this configuration. This is
expected and not informative: the native class head is not trained in the
sequential setting and operates at chance level, so its predictions do not
reflect the concept-based decision pathway.

For CEM, the nearest-code classification accuracy is substantially higher
($70.50\%$), reflecting that its concept outputs are strongly class-informative.
However, the native model still achieves higher accuracy than both the
codebook-based prediction and the discrete-concept probe, resulting in a
non-zero gap (Table~\ref{tab:leakage_probe}). This indicates that, despite
the informativeness of the concept representation, the final prediction is not
fully determined by the exposed concept channel.

\paragraph{Interpretation.}
Taken together, these results provide a complementary view of leakage.
While probe-based analysis reveals whether predictions are reconstructible from
concept outputs, codebook fidelity evaluates whether predictions follow a
discrete concept-level decision rule. The fully constrained configuration satisfies both criteria,
whereas CEM exhibits residual performance beyond what can be explained by
concept-based reasoning, consistent with the leakage gaps observed in the main
text.


\section{Common Implementation and Training Details}\label{app:training}

\subsection{Ablation Template and Fully Constrained Configuration}
\label{app:template}
\begin{figure}[h]
    \centering
    \includegraphics[width=\linewidth]{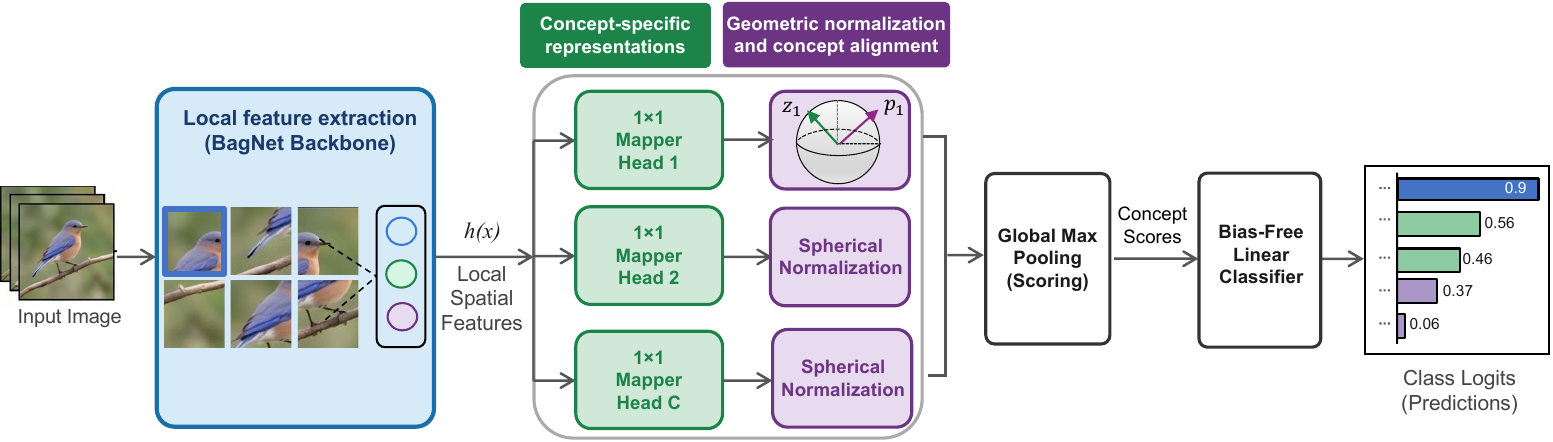}
    \caption{Fully constrained configuration of the ablation template
    (\modelname{}). A BagNet-33 backbone extracts local features, projects
    them into concept-specific representations, compares them with learned
    concept vectors via cosine similarity, and aggregates them via global
    max pooling into concept scores. The other configurations of
    Table~\ref{tab:ablations} replace the backbone, the heads, the scoring,
    or the training objective.}
    \label{fig:architecture}
\end{figure}

We describe the template in its fully constrained configuration; the
ablated configurations change one component at a time (backbone:
ResNet50 or BagNet-$k$; heads: one shared head instead of grouped
per-concept projections; scoring: dot product instead of cosine
similarity; training: joint instead of concept-only). The fully
constrained configuration processes an input image through a
five-stage pipeline:
\begin{itemize}
    \item \textbf{BagNet-33 backbone}: produces spatial feature maps
        with a hard $33\times33$ pixel receptive field per position,
        yielding a tensor of shape $(B, C, H, W)$.
    \item \textbf{Latent mapper}: a stack of $1\times1$ convolutions
        projects backbone features into per-concept latent subspaces.
        The first stage is a dense projection; subsequent stages use
        grouped convolutions with $n$ groups so that each concept
        operates on its own $d$-dimensional channel slice.
        Each spatial position is L2-normalized per concept, yielding
        latents of shape $(B, n, d, H, W)$ on the unit sphere.
    \item \textbf{Prototype similarity maps}: $l$ prototype vectors
        per concept (each of dimension $d$) are compared via cosine
        similarity at every spatial location using grouped convolution,
        yielding similarity maps of shape $(B, n, l, H, W)$ with
        values in $[-1, 1]$.
    \item \textbf{Global max pooling}: spatial and prototype dimensions
        are collapsed by taking the maximum over all $(H \times W)$
        locations and over all $l$ prototypes per concept, producing
        one continuous concept score per concept: $s \in \mathbb{R}^n$.
    \item \textbf{(Optional) Linear classification head}
        $\mathbb{R}^n \to \mathbb{R}^{n_\text{classes}}$ maps concept
        scores to class logits.
\end{itemize}

Discretization of $s$ into $\{0, 1\}^n$ is applied
post-hoc during evaluation and does not participate in gradient
updates.

\paragraph{Training Procedure}
All main-text \modelname{} results use \emph{concept-only} training: the
linear classification head is frozen (never updated) and the backbone,
latent mapper, and prototypes are optimized with Adam using only the
concept loss,
\[
\mathcal{L} = \mathcal{L}_\text{concept},
\]
where $\mathcal{L}_\text{concept}$ is the weighted binary cross-entropy of
Eq.~\eqref{EQ:Weighted-BEC-Concepts} (certainty weights for
\textit{by-instance} labels, unit weights for \textit{by-class} labels).
Model checkpointing uses best validation concept accuracy. Because the head
is never trained under this protocol, its own predictions are at chance
level; task accuracy for \modelname{} is instead obtained by the post-hoc
linear probe of Section~\ref{par:leakage_probe}, fit on frozen concept
scores exported from the checkpoint. This is also why
$\Delta_{\mathrm{bypass}}=0$ holds by construction for \modelname{}: the
only path from concept scores to a class prediction is the probe itself.

The \emph{joint-training} ablation reported in
Table~\ref{tab:ablations} (\S\ref{par:ablation}) instead optimizes all
parameters end-to-end every epoch, including the class head, with
\[
\mathcal{L} = \mathcal{L}_\text{CE}(\hat{y}, y)
            + \lambda \cdot \mathcal{L}_\text{concept},
\]
where $\mathcal{L}_\text{CE}$ is cross-entropy on class logits. This
variant is used only to test whether task supervision changes concept
scores (\S\ref{par:ablation}); it is not the protocol behind the main
\modelname{} rows elsewhere in the paper.

\paragraph{Hyperparameters}
All reported \modelname{} results use the configuration in Table~\ref{tab:appendix_hyperparam}. 

\begin{table}[h]
\centering
\small
\caption{\modelname{} hyperparameters used for all reported experiments.}
\begin{tabular}{ll}
\toprule
Hyperparameter & Value \\
\midrule
epochs & 150 \\
learning\_rate & $1 \times 10^{-4}$ \\
weight\_decay & $1 \times 10^{-4}$ \\
batch\_size & 32 \\
image\_size & $224 \times 224$ \\
bbox\_center\_crop & True \\
latent\_out\_channels & 128, 64, 32 \\
bagnet\_latent\_out\_channels & 1024, 256 \\
n\_prototypes\_per\_concept & 1 \\
optimizer & Adam \\
lr\_scheduler & 
\begin{tabular}[c]{@{}l@{}}ReduceLROnPlateau\\ (factor=0.05, patience=10, min\_lr=$10^{-5}$)\end{tabular}
\\
\bottomrule
\end{tabular}
\label{tab:appendix_hyperparam}
\end{table}

Weight decay is applied to backbone and mapper parameters only; the prototype weight tensor receives no decay.

\paragraph{Augmentation and Pre-processing: \modelname{} vs.\ CEM}
\textbf{\modelname{}} uses the CUB bounding box annotation (\texttt{bbox\_center\_crop=True}) to centre-crop each image around the annotated bird bounding box before resizing to $224 \times 224$. This preserves the bird's aspect ratio and ensures that patch-level features at every spatial position correspond to predictable, bird-relative image regions. Shape and size attributes, which are aspect-ratio-sensitive, are not distorted, and part-level attributes map to consistent spatial positions across images.

\textbf{CEM (and standard CBM)} uses a standard ImageNet-style preprocessing pipeline (resize + random crop) that does not account for the bounding box. This pipeline can significantly distort the bird's aspect ratio, shift the bird within the crop, and truncate body parts. Shape and size attributes are confounded by crop geometry, and spatial precision for all part-level attributes is reduced because the relationship between pixel position and bird anatomy is not consistent across images.

This preprocessing difference is a key experimental variable: \modelname{}'s spatial faithfulness results (Appendix~\ref{app:spatial_faith}) are only interpretable under \texttt{bbox\_center\_crop=True}.

\subsection{Baselines: CEM and Joint CBM}

\textbf{CEM}~\cite{espinosa2022concept} is trained using the authors' published implementation with a ResNet-18 feature extractor, embedding size 16, \texttt{concept\_loss\_weight=1}, and \texttt{training\_intervention\_prob=0.25}. The optimizer is SGD (lr=0.01, momentum=0.9, weight\_decay=$4 \times 10^{-6}$) with early stopping (patience=15) up to 150 epochs.

\textbf{Joint CBM}~\cite{koh2020concept} is trained end-to-end with \texttt{concept\_loss\_weight=3} using the same ResNet-18 backbone and preprocessing pipeline as CEM.

Both baselines are trained with 3 independent seeds (42, 43, 44). For the \textbf{\textit{by-class} evaluation}, all three seeds use Koh et al.'s class-level concept labels. For the \textbf{\textit{by-instance} evaluation}, the same three seeds are retrained on per-image CUB instance annotations. Val/test splits follow the standard Koh et al.\ preprocessing (4,796 train / 1,000 val / 5,794 test).

\begin{table}[h]
\centering
\small
\caption{Concept-to-CUB-part mapping used in the C5 spatial faithfulness evaluation. Concepts \texttt{has\_size\_*}, \texttt{has\_shape\_*}, and \texttt{has\_primary\_color\_*} are excluded (no anatomical anchor).}
\begin{tabular}{lll}
\toprule
Concept prefix & CUB part IDs & Bilateral? \\
\midrule
\texttt{bill\_*} & beak (2) & No \\
\texttt{belly\_*} & belly (3) & No \\
\texttt{breast\_*} & breast (4) & No \\
\texttt{crown\_*} & crown (5) & No \\
\texttt{forehead\_*} & forehead (6) & No \\
\texttt{eye\_color} & left eye (7), right eye (11) & Yes \\
\texttt{leg\_*} & left leg (8), right leg (12) & Yes \\
\texttt{wing\_*} & left wing (9), right wing (13) & Yes \\
\texttt{nape\_*} & nape (10) & No \\
\texttt{tail\_*} & tail (14) & No \\
\texttt{throat\_*} & throat (15) & No \\
\texttt{back\_*} & back (1) & No \\
\texttt{head\_pattern} & crown (5), forehead (6), nape (10) & No \\
\texttt{underparts\_*} & belly (3), breast (4) & No \\
\texttt{upperparts\_*} & back (1), crown (5) & No \\
\texttt{under\_tail\_*} & tail (14), belly (3) & No \\
\texttt{upper\_tail\_*} & tail (14), back (1) & No \\
\bottomrule
\end{tabular}
\label{tab:appendix_concept_to_cub}
\end{table}

\subsection{Concept-to-CUB-Part Mapping}
\label{app:concept_cub_mapping}
The Spatial faithfulness evaluation requires mapping each CUB concept to one or more anatomical keypoints. The canonical mapping covers 17 anatomical groups and is shown in Table~\ref{tab:appendix_concept_to_cub} and stored in \texttt{notebooks/concept\_to\_cub\_part\_mapping.json}.
Three concept families are \textbf{excluded from the C5 evaluation}: \texttt{has\_size\_*}, \texttt{has\_shape\_*}, and \texttt{has\_primary\_color\_*}. These are global descriptors with no anatomical part anchor. For \textbf{bilateral parts} (eyes, legs, wings), the nearest visible keypoint is used as the evaluation target rather than the mean of left and right keypoints.

\section{Cross-Label Concept Accuracy}\label{app:cross_label}

This appendix reports concept accuracy when models are evaluated against a label regime different from their training regime, establishing that Koh (class-wise) and instance-wise annotations are substantially consistent.

\paragraph{Setup.} Koh-trained model scores are evaluated against instance-wise CUB ground truth; instance-trained model scores are evaluated against Koh class-level labels. Three model families (\modelname{}, CEM, CBM) are evaluated across three data splits each.

\begin{table}[h]
\centering
\small
\begin{tabular}{lcccc}
\toprule
Family &
  \begin{tabular}[c]{@{}c@{}}Koh-trained $\to$ \\ Instance GT\end{tabular} &
  \begin{tabular}[c]{@{}c@{}}Instance-trained  $\to$ \\ Koh GT\end{tabular} &
  \begin{tabular}[c]{@{}c@{}}Koh$\to$ \\ Koh (matched)\end{tabular} &
  \begin{tabular}[c]{@{}c@{}}Instance  $\to$ \\ Instance (matched)\end{tabular} \\ 
\midrule
\modelname{} & 0.800 & 0.777 & 0.904 & 0.748 \\
CEM            & 0.812 & 0.841 & 0.949 & 0.798 \\
CBM            & 0.811 & 0.841 & 0.946 & 0.800 \\
\bottomrule
\end{tabular}
\caption{Cross-label concept accuracy (mean across 3 splits). Matched-label baselines are shown for reference only.}
\label{tab:cross_label}
\end{table}

\paragraph{Interpretation.}
\begin{itemize}
    \item \textbf{Label regimes are substantially consistent.} A model trained exclusively on Koh labels reaches $80$--$84\%$ accuracy when evaluated on instance-wise CUB GT, and vice versa. This confirms that both annotation protocols capture the same underlying bird attributes despite differing in granularity and derivation.
    \item \textbf{\modelname{} generalizes less across label regimes than CEM and CBM} ($-1$ to $-6$ pp). The concept representations learned by \modelname{} are more annotation-regime-dependent, particularly in the instance-trained $\to$ Koh GT direction ($-6.4$ pp). This is consistent with the instance-trained \modelname{} being a weaker concept predictor overall (matched accuracy $74.8\%$ vs.\ $79.8$--$80.0\%$ for CEM/CBM).
    \item \textbf{For CEM and CBM, instance-trained $\to$ Koh GT} ($\approx$84\%) outperforms Koh-trained $\to$ instance GT ($\approx$81\%) by $\approx$3 pp. Models trained on noisier instance-wise labels still recover the Koh-measurable concept signal at high accuracy, supporting instance-wise training as a more faithful annotation protocol without sacrificing Koh-measurable concept quality.
\end{itemize}

This analysis confirms that the two label sets are not measuring incompatible things, and that results under different labeling regimes are directly comparable. The $\approx$18\% cross-label disagreement rate bounds the expected concept accuracy of any faithfully grounded model, as discussed in Section~\ref{sec:ceilings}.

\begin{figure*}[t]
\centering

\textbf{Class-level (Koh)}\\[2mm]

\begin{subfigure}{0.32\linewidth}
    \centering
    \includegraphics[width=\linewidth]{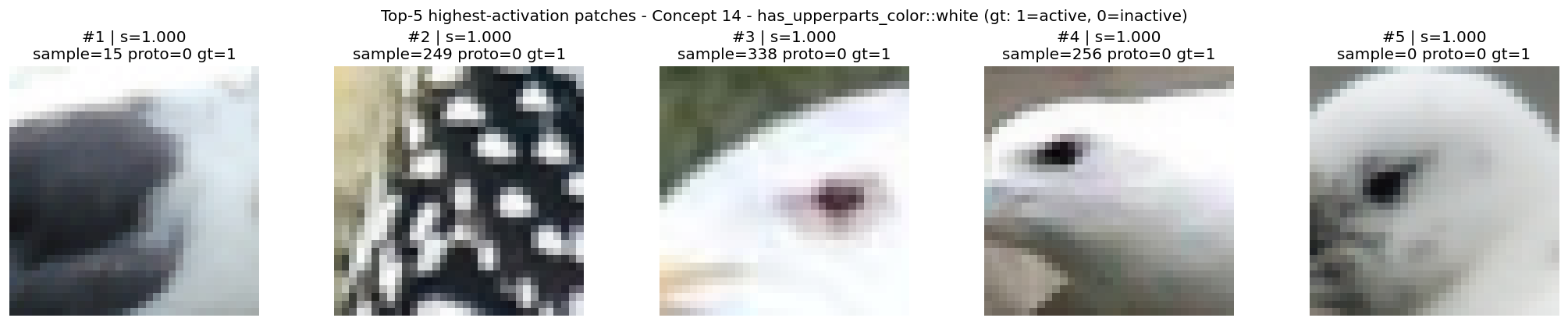}
    \caption{Concept 14}
\end{subfigure}
\hfill
\begin{subfigure}{0.32\linewidth}
    \centering
    \includegraphics[width=\linewidth]{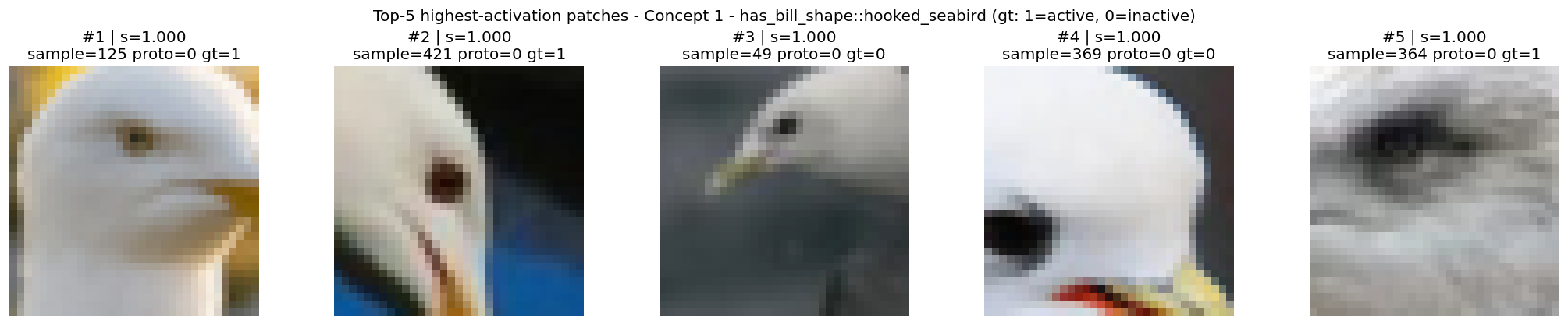}
    \caption{Concept 1}
\end{subfigure}
\hfill
\begin{subfigure}{0.32\linewidth}
    \centering
    \includegraphics[width=\linewidth]{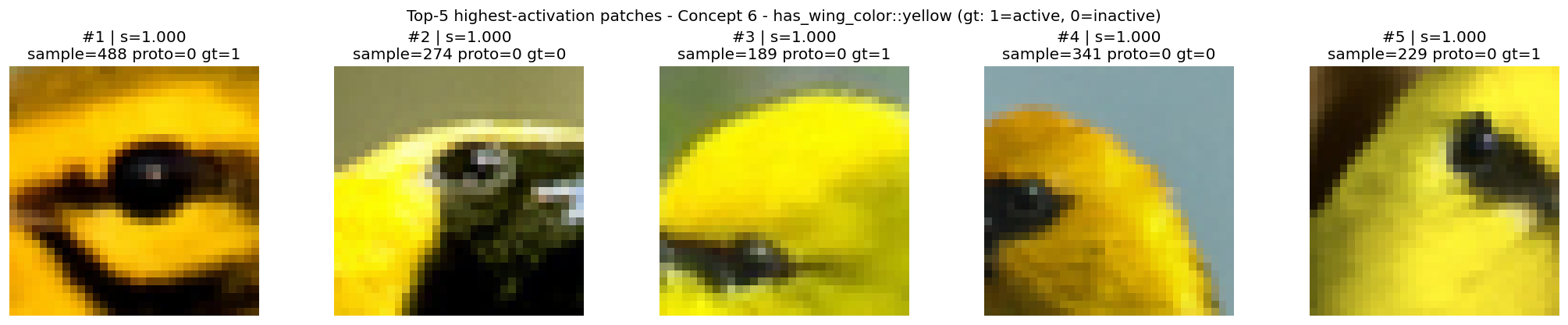}
    \caption{Concept 6}
\end{subfigure}

\vspace{4mm}

\textbf{Instance-level}\\[2mm]

\begin{subfigure}{0.32\linewidth}
    \centering
    \includegraphics[width=\linewidth]{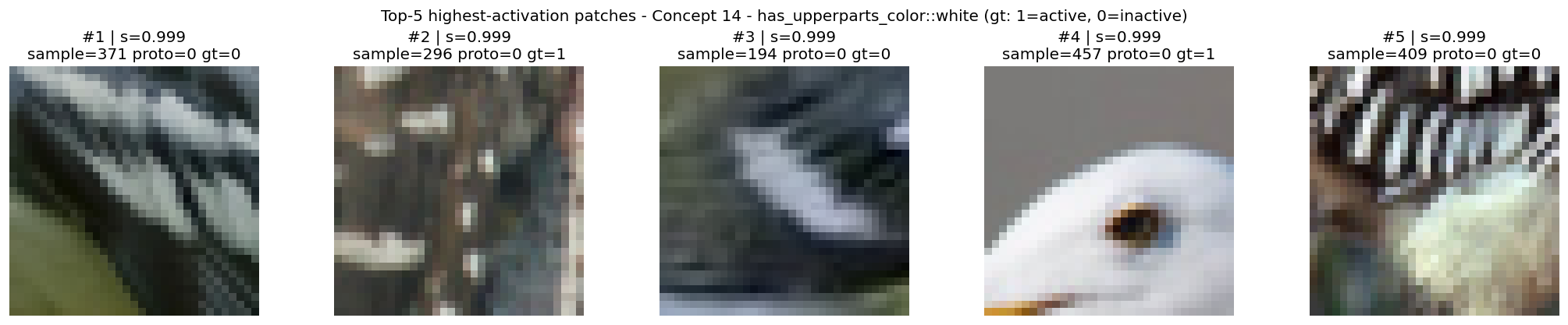}
    \caption{Concept 14}
\end{subfigure}
\hfill
\begin{subfigure}{0.32\linewidth}
    \centering
    \includegraphics[width=\linewidth]{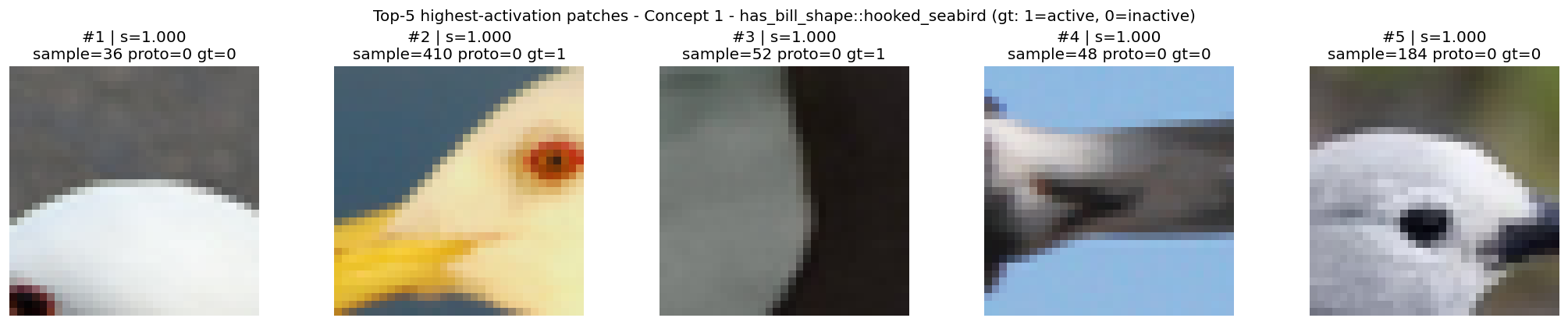}
    \caption{Concept 1}
\end{subfigure}
\hfill
\begin{subfigure}{0.32\linewidth}
    \centering
    \includegraphics[width=\linewidth]{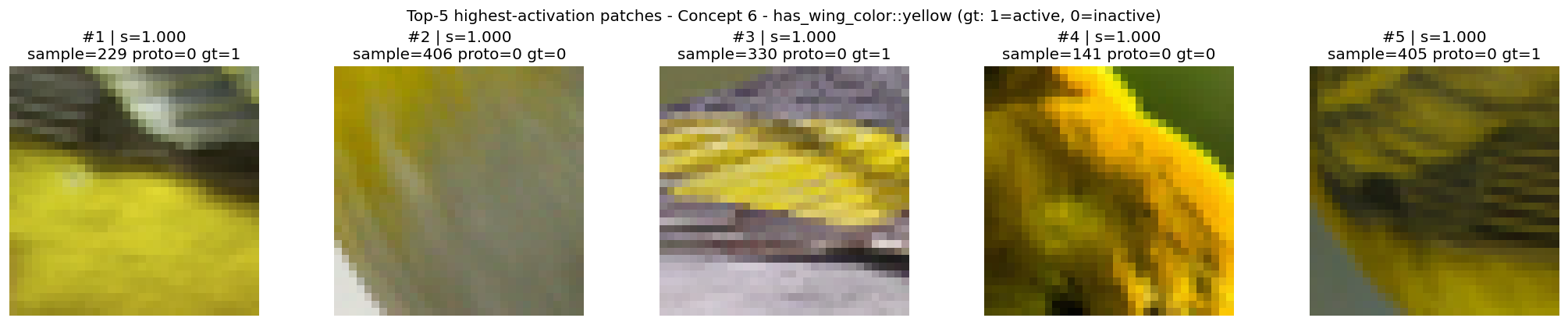}
    \caption{Concept 6}
\end{subfigure}

\caption{Top-5 highest-activation patches for selected concepts under class-level (top) and instance-level (bottom) supervision. 
Class-level training shows a strong bias toward head regions, while instance-level supervision produces more spatially aligned activations.}
\label{fig:proto_compare}
\end{figure*}

\section{Qualitative Visualisations}\label{app:visualisations}

\subsection{Concept Representation}
We provide qualitative evidence of concept grounding by visualizing top-activating patches for a small set of representative concepts. 
For each concept, we show the top-5 highest-activation patches, corresponding to the spatial locations that most strongly support the prediction.

\paragraph{Selection.}
We select concepts that highlight different behaviors: 
(i) \texttt{has\_upperparts\_color::white} (global color), 
(ii) \texttt{has\_bill\_shape::hooked\_seabird} (localized shape), and 
(iii) \texttt{has\_wing\_color::yellow} (mid-level attribute). 
These examples illustrate both success and failure modes.

\paragraph{Results.}
Figure~\ref{fig:proto_compare} compares class-level (Koh) and instance-level supervision.

Under class-level supervision, activations are often concentrated on head regions, even for concepts unrelated to the head, indicating reliance on class-correlated cues. 
In contrast, instance-level supervision yields more spatially aligned activations, particularly for localized concepts such as bill shape and wing color.

\paragraph{Interpretation.}
These visualizations support the quantitative findings: leakage manifests as a reuse of shared discriminative regions, while instance-level supervision encourages concept-specific, spatially grounded representations.

\subsection{Overlay of Cub key point and BagNet RF}

\begin{figure}[h]
\centering

\begin{subfigure}{0.45\textwidth}
    \includegraphics[width=\linewidth]{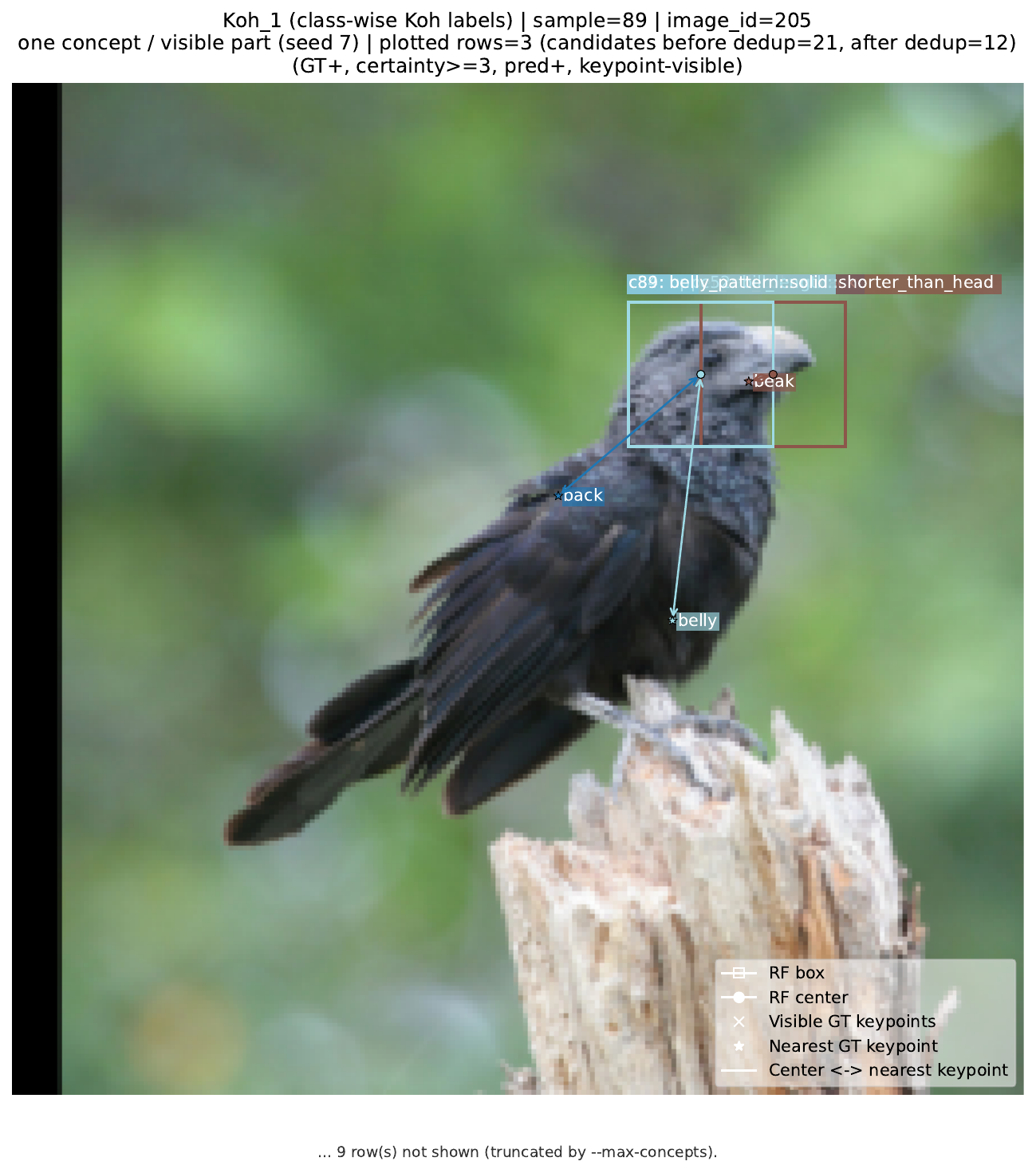}
    \caption{Koh (by class), idx 000089}
\end{subfigure}
\hfill
\begin{subfigure}{0.45\textwidth}
    \includegraphics[width=\linewidth]{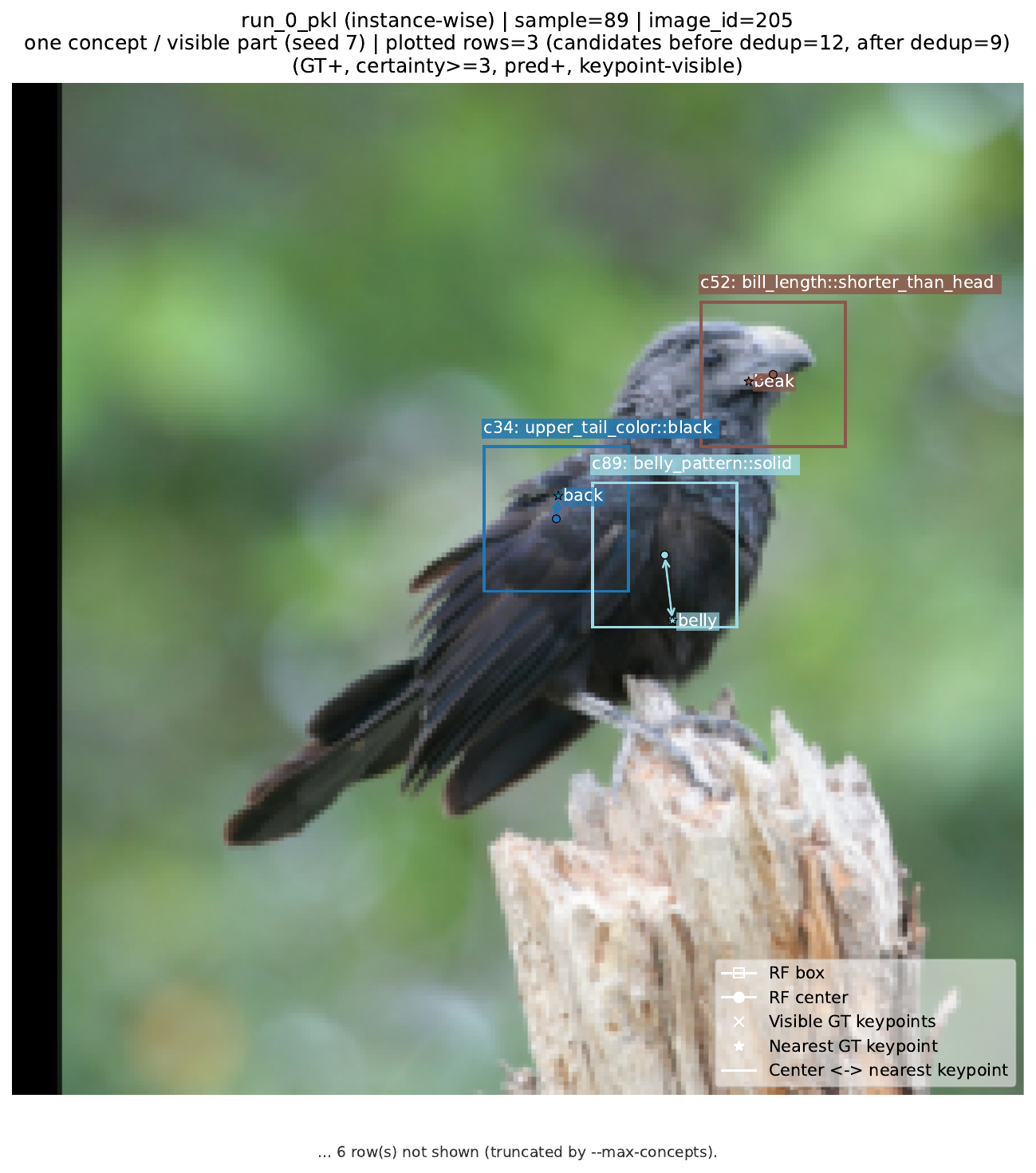}
    \caption{Instance-wise, idx 000089}
\end{subfigure}

\vspace{0.4cm}

\begin{subfigure}{0.45\textwidth}
    \includegraphics[width=\linewidth]{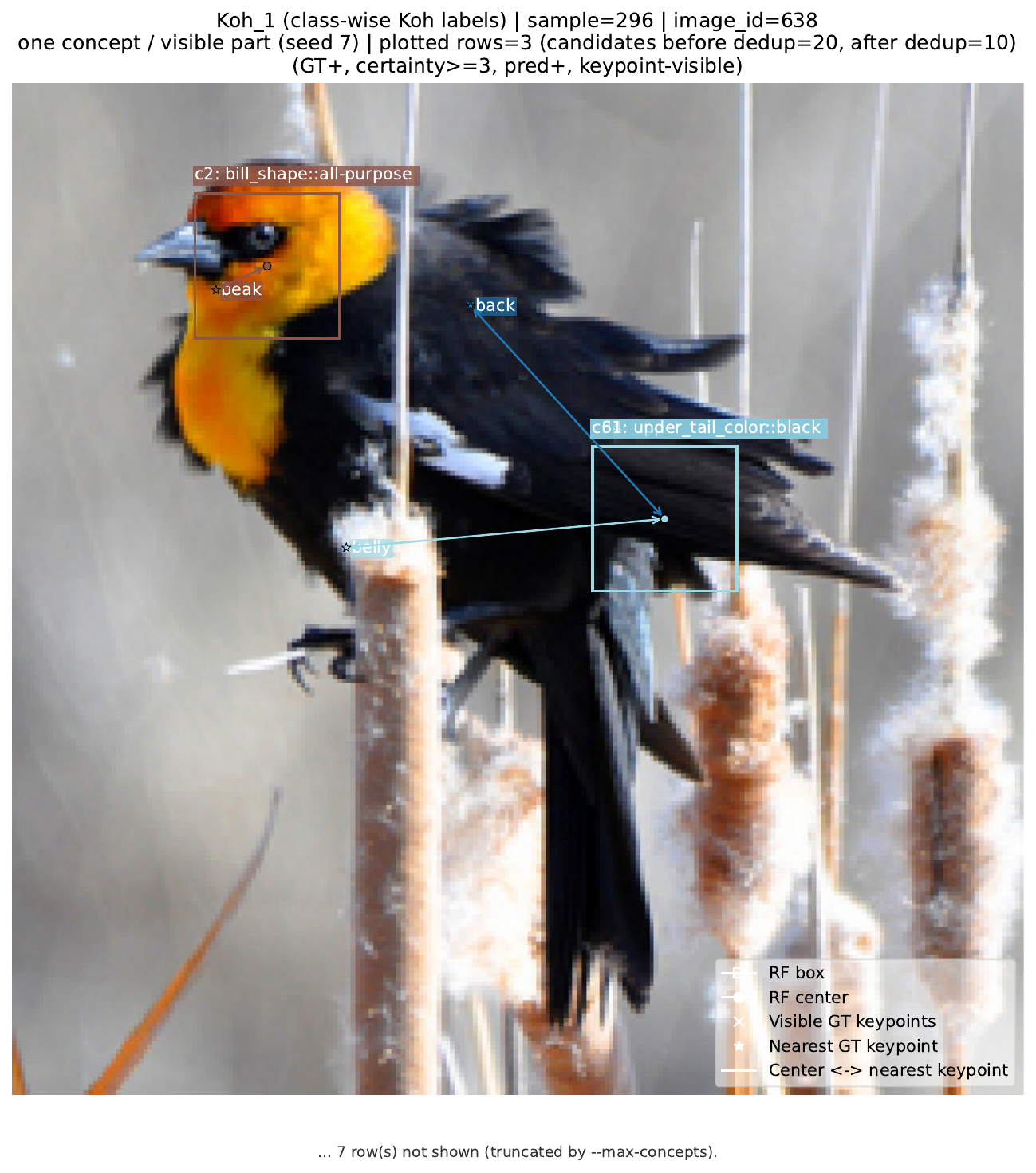}
    \caption{Koh (by class), idx 000296}
\end{subfigure}
\hfill
\begin{subfigure}{0.45\textwidth}
    \includegraphics[width=\linewidth]{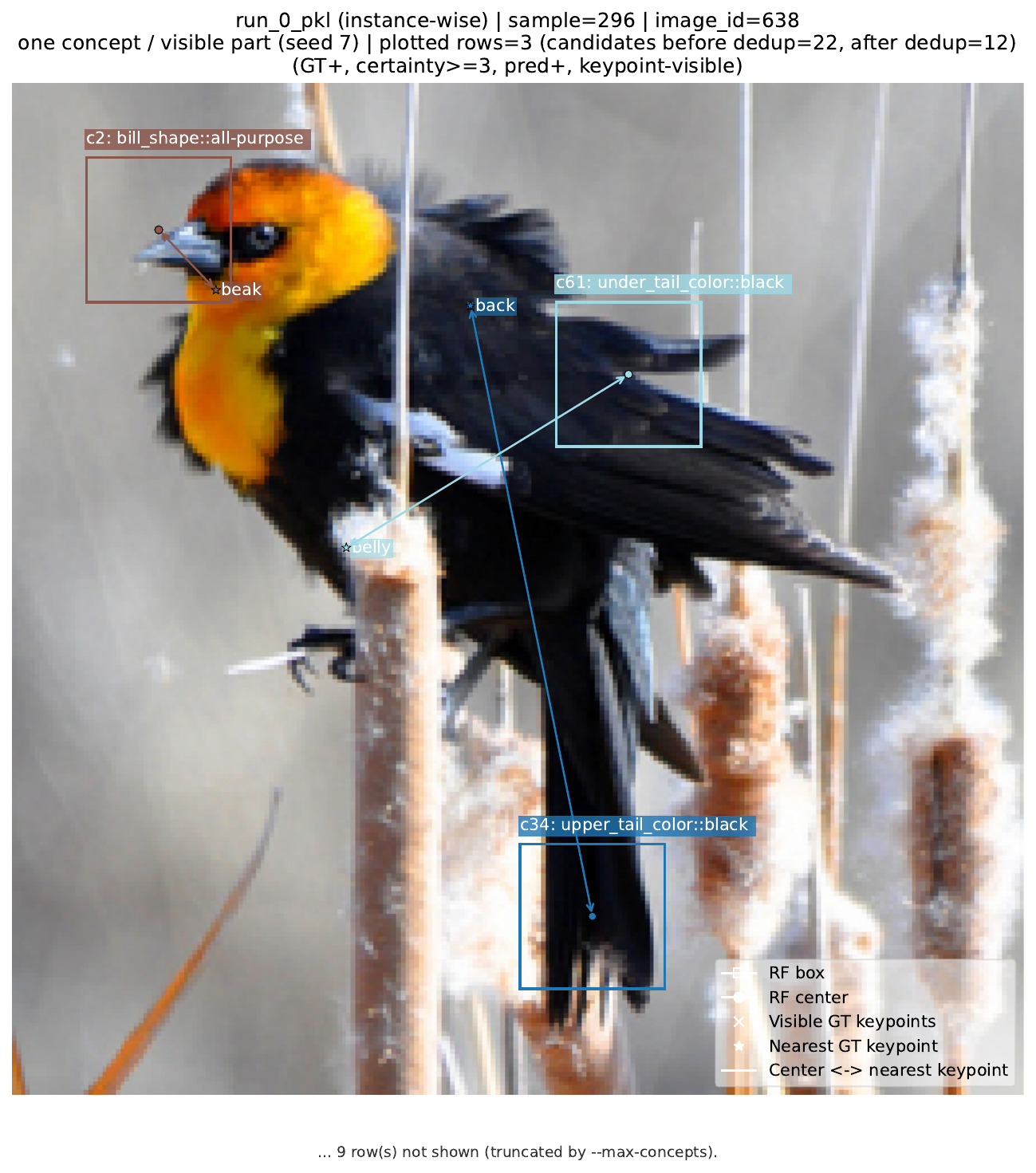}
    \caption{Instance-wise, idx 000296}
\end{subfigure}

\caption{Qualitative visualization of concept localization. For each sample, we show the keypoint (center) and the corresponding receptive field (bounding box) used by the network to infer the concept. We compare class-level (Koh) and instance-level supervision across two examples. Instance-level supervision yields activations that are more spatially aligned with the expected anatomical regions.}
\label{img:qualitative:overlay}
\end{figure}

Figure~\ref{img:qualitative:overlay} visualizes both the keypoint provided by the original CUB dataset, and the corresponding receptive field, illustrating how the model grounds concepts in local image evidence. Consistent with our spatial faithfulness analysis, instance-level supervision leads to better alignment between receptive fields and anatomical keypoints, whereas class-level supervision often relies on less localized, proxy regions.

\end{document}